\documentclass{article}

\PassOptionsToPackage{numbers, compress}{natbib}
\usepackage[final]{neurips_2019}

\usepackage[utf8]{inputenc} % allow utf-8 input
\usepackage[T1]{fontenc}    % use 8-bit T1 fonts
\usepackage{hyperref}       % hyperlinks
\usepackage{url}            % simple URL typesetting
\usepackage{booktabs}       % professional-quality tables
\usepackage{amsfonts}       % blackboard math symbols
\usepackage{nicefrac}       % compact symbols for 1/2, etc.
\usepackage{microtype}      % microtypography

\usepackage{soul}
\usepackage{graphicx}
\usepackage{amsmath}
\usepackage{amssymb}
\usepackage{multirow}
\usepackage{mathtools}
\usepackage[normalem]{ulem}
\usepackage{tabu}
\usepackage{scientific_abbrev}
\usepackage{xfrac}
\usepackage[ruled,linesnumbered]{algorithm2e}
\usepackage{float} 
\usepackage{overpic}
\usepackage{cuted}
\usepackage{capt-of}
\usepackage{url}
\usepackage{enumitem}
\usepackage{threeparttablex, tablefootnote}

\DeclareMathOperator{\atantwo}{atan2}
\newcommand{\twodots}{\mathinner {\ldotp \ldotp}}

\usepackage{overpic}
\usepackage{tabulary}
\usepackage{color}
\usepackage{gensymb}

\newcommand{\comment}[1]{}
\definecolor{orange}{RGB}{255,127,0}
\definecolor{green}{RGB}{34,139,34}

\DeclarePairedDelimiter\floor{\lfloor}{\rfloor}
\DeclareMathOperator{\SSIM}{SSIM}
\DeclareMathOperator{\lstm}{LSTM}

\renewcommand{\paragraph}[1]{\par\vspace{0.2em}\noindent{\bf #1}}

\newcommand{\beginsupplement}{%
        \setcounter{table}{0}
        \renewcommand{\thetable}{S\arabic{table}}%
        \setcounter{figure}{0}
        \renewcommand{\thefigure}{S\arabic{figure}}%
     }

\makeatletter
\newcommand\refwithdefault[2]{%
  \@ifundefined{r@#1}{%
    #2%
  }{%
    \ref{#1}%
  }%
}
\makeatother

\graphicspath {{./figures/}}

\definecolor{citecolor}{RGB}{34,139,34}

\title{Incremental Scene Synthesis}

\author{\textbf{Benjamin Planche}$^{1,2}$\qquad
		\textbf{Xuejian Rong}$^{3,4}$\qquad
		\textbf{Ziyan Wu}$^{4}$\qquad
		\textbf{Srikrishna Karanam}$^{4}$ \\[.3em]
		\textbf{Harald Kosch}$^{2}$\qquad
		\textbf{YingLi Tian}$^{3}$\qquad
		\textbf{Jan Ernst}$^{4}$\qquad 
		\textbf{Andreas Hutter}$^{1}$
\vspace{0.5em}\\
$^{1}$Siemens Corporate Technology, Munich, Germany\\
$^{2}$University of Passau, Passau, Germany\\
$^{3}$The City College, City University of New York, New York NY\\
$^{4}$Siemens Corporate Technology, Princeton NJ\\\vspace{0.3em}
{\tt\small \{first.last\}@siemens.com, \{xrong,ytian\}@ccny.cuny.edu, harald.kosch@uni-passau.de}
\vspace{-0.9em}
}

\begin{document}

\maketitle

\begin{abstract}
We present a method to incrementally generate complete 2D or 3D scenes with the following properties: (a) it is globally consistent at each step according to a learned scene prior, (b) real observations of a scene can be incorporated while observing global consistency, (c) unobserved regions can be hallucinated locally in consistence with previous observations, hallucinations and global priors, and (d) hallucinations are statistical in nature, \ie, different scenes can be generated from the same observations. To achieve this, we model the virtual scene, where an active agent at each step can either perceive an observed part of the scene or generate a local hallucination. The latter can be interpreted as the agent's expectation at this step through the scene and can  be applied to autonomous navigation. In the limit of observing real data at each point, our method converges to solving the SLAM problem. It can otherwise sample entirely imagined scenes from prior distributions. Besides autonomous agents, applications include problems where large data is required for building robust real-world applications, but few samples are available. We demonstrate efficacy on various 2D as well as 3D data.
\end{abstract}

% Added by SK for teaser fig, please comment until end of teaser fig if this is not required
\begin{figure}[h!]
    \centering
    \includegraphics[width=1\linewidth]{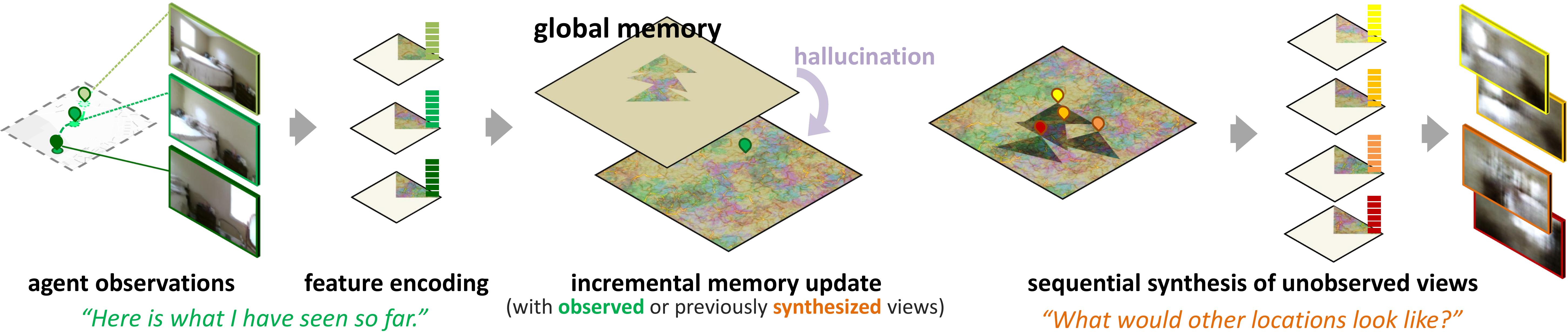}
    \vspace{-1.5em}
	\captionof{figure}{\textbf{Our solution} for scene understanding and novel view synthesis, given non-localized agents.}
	% \textbf{Our solution} to register observations from a non-localized agent in a global feature map and generate novel views accordingly.}
	\label{fig:teaser}
% 	\vspace{1.5em}
\end{figure}
% \twocolumn[{%
% \renewcommand\twocolumn[1][]{#1}%
% \maketitle
% \vspace{-6.5em}
% \begin{center}
%     \centering
%     %\includegraphics[scale=0.8]{teaser1_v2.png}
%     \includegraphics[width=1\linewidth]{pipeline_with_avd_colors_flat_new2}
%     \vspace{-1.7em}
%     \captionof{figure}{\textbf{Our solution} to register observations from a non-localized agent in a global feature map and generate novel views accordingly.}
%     \label{fig:pipeline}
% \end{center}%
% }]
%%%%end teaser fig%%%%

\section{Introduction}
%\currentpdfbookmark{Introduction}{Introduction}
% !TeX spellcheck = en_US

%\begin{figure*}[t]
%\centering
%\includegraphics[width=1\linewidth]{pipeline_prev}
%\caption{\textbf{Overall Pipeline}. The proposed system is able to register observations from %a non-localized agent in a global feature, which can then be used to generate new %observations.} %\cBen{todo: adapt terms and add formal notations for all figures}}
%\label{fig:pipeline}  
%\end{figure*}

We live in a three-dimensional world, and a proper cognitive understanding of its structure is crucial for planning and action.
The ability to anticipate under uncertainty is necessary for autonomous agents to perform various downstream tasks such as  exploration and target navigation~\citep{singh2018active}. 
Deep learning has shown promise in addressing these questions~\citep{zhou2016view,park2017transformation}.
% especially on the \textit{Novel view synthesis} task, which focuses on synthesizing the observation of an autonomous agent from a novel viewpoint, given one or more previous observations.
Given a set of views and corresponding camera poses, existing methods have demonstrated the capability of learning an object's 3D shape via direct 3D or 2D supervision.
% the reprojection error: given an estimated shape, one can project it to the known camera views and compare to the provided images. The discrepancy between these generated projections and training samples provides supervisory signal for improving the shape estimate.

\textsl{Novel view synthesis} methods of this type have three common limitations.
First, most recent approaches solely focus on single objects and surrounding viewpoints, and are trained with category-dependent 3D shape representations (\eg, voxel, mesh, point cloud model) and 3D/2D supervision (\eg, reprojection loss), which are not trivial to obtain for natural scenes.
While recent works on auto-regressive pixel generation \citep{tatarchenko2016multi}, appearance flow prediction \citep{zhou2016view}, or a combination of both \citep{sun2018multi} generate encouraging preliminary results for scenes, they only evaluate on data with mostly forwarding translation (\eg, KITTI dataset~\citep{Geiger2012CVPR}), and no scene understanding capabilities are convincingly shown.
% Second, most methods do not support a varying number of conditioning inputs.
Second, these approaches assume that the camera poses are known precisely for all provided observations. This is a practically and biologically unrealistic assumption; an agent typically only has access to its own observations, not its precise location relative to objects in the scene (albeit it is provided by some oracle in synthetic environments, \eg, \citep{eslami2018neural}).
Third, there are no constraints to guarantee consistency among the synthesized results.

In this paper, we address these issues with a unified framework that incrementally generates complete 2D or 3D scenes (\cf Figure~\ref{fig:teaser}). Our solution builds upon the MapNet system~\citep{henriques2018mapnet}, which offers an elegant solution to the registration problem but has no memory-reading capability. In comparison, our method not only provides a completely functional memory system, but also displays superior generation performance when compared to parallel deep reinforcement learning methods (\eg, \citep{fraccaro2018generative}). To the best of our knowledge, our solution is the first complete end-to-end trainable read/write allocentric spatial memory for visual inputs. % It not only surpasses recent deep SLAM approaches (e.g.,  \cite{tateno2017cnnslam}) with prediction and synthesizing capabilities, and also shows superior generation performance than parallel deep reinforcement learning methods (e.g., \cite{fraccaro2018generative}).
Our key contributions are summarized below:

%. We begin with non-localized active agents, modeling its motion through a virtual scene. As the agent moves through the scene, it only provides us with observations (e.g., images depicting observed part of the scene), and no location information. We then use these observations to compute features and perform registration, which allows us to localize the agent. At this point, we also begin to construct and update a global representation of the scene given    

%To summarize, the key contributions of this paper include:
\begin{itemize}[leftmargin=*]
    %\item Starting with only scene observations provided by an active, non-localized agent, we present a mechanism to build a global representation of the scene while also localizing the agent itself.
    %\item We present novel mechanisms to both use the global representation to hallucinate unobserved parts of the scene as well as update the global representation with both observed and hallucinated data.
    \item Starting with only scene observations from a non-localized agent (\ie, no location/action inputs unlike, \eg, 
    %\cite{fraccaro2018generative,zhangneural,rosenbaum2018learning}
    \citep{fraccaro2018generative}), we present novel mechanisms to update a global memory with encoded features, hallucinate unobserved regions and query the memory for novel view synthesis.
    \item Memory updates are done with either observed or hallucinated data. Our domain-aware mechanism is the first to explicitly ensure the representation's global consistency \wrt the underlying scene properties in both cases.
    \item We propose the first framework that integrates observation, localization, globally consistent scene learning, and hallucination-aware representation updating to enable incremental scene synthesis.
\end{itemize}

We demonstrate the efficacy of our framework on a variety of partially observable synthetic and realistic 2D environments. Finally, to establish scalability, we also evaluate the proposed model on challenging 3D environments.

%-------------------------------------------------------------------------
\section{Related Work}
\label{sec:rw}
    % !TeX spellcheck = en_US
    
%Localization and mapping are long-standing problems in computer vision, constituting the field of 3D robotics and navigation. Video prediction and novel view synthesis have also gained significant interest. Related to both research areas, we discuss them below to put our work in context.

\begin{figure}[t]
    \centering
    \includegraphics[width=1\linewidth]{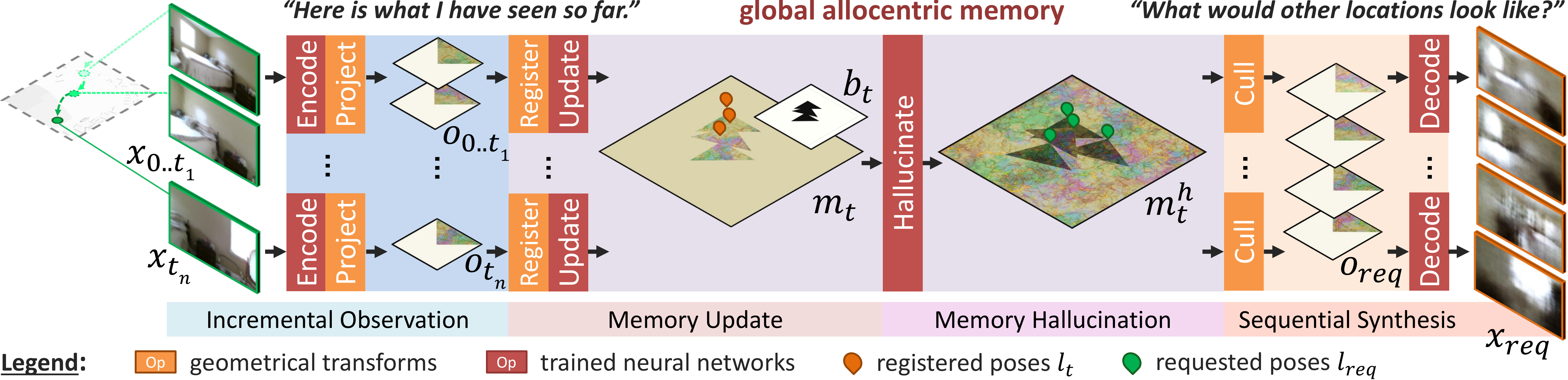}
    \vspace{-1.5em}
	\captionof{figure}{\textbf{Proposed pipeline} for non-localized agents exploring new scenes. Observations $x_t$ are sequentially encoded and registered in a global feature map $m_t$ with spatial properties, used to extrapolate unobserved content and generate consistent novel views $x_{req}$ from requested viewpoints.}
	% \textbf{Our solution} to register observations from a non-localized agent in a global feature map and generate novel views accordingly.}
	\label{fig:pipeline}
% 	\vspace{1.5em}
\end{figure}

Our work is related to localization, mapping, and novel view synthesis. We discuss relevant work to provide some context.
    
\paragraph{Neural Localization and Mapping.} The ability to build a global representation of an environment, by registering frames captured from different viewpoints, is key to several concepts such as reinforcement learning or scene reconstruction. 
%The goal is to fill a global structure so registered content can be easily retrieved, and new views \eg, from different viewpoints, can be interpolated or rendered accordingly. 
Recurrent neural networks are commonly used to accumulate features from image sequences, \eg, to predict the camera trajectory~\citep{parisotto2018global,rosenbaum2018learning}.
Extending these solutions with a queryable memory, state-of-the-art models are mostly egocentric and action-conditioned~\citep{singh2018active,pritzel2017neural,zhangneural,fraccaro2018generative,parisotto2018neuralmap}.
%They usually consider an agent exploring the environment, which can provide not only an observation $x_t$ at each time step $t$, but also its state $s_t$ (\eg, its pose) or action $a_t$ leading to this new observation (\eg ``move forward'', ``rotate clockwise'',  \etc).
Some oracle is, therefore, usually required to provide the agent's action at each time step $t$~\citep{parisotto2018neuralmap}.
This information is typically used to regress the agent state $s_t$, \eg, its pose, which can be used in a memory structure to index the corresponding observation $x_t$ or its features. In comparison, our method solely relies on the observations to regress the agent's pose.  %\cSK{Perhaps add one line on how we are different.}\cBen{done, really briefly} \cSK{Looks good, I'll remove the comment then.}
%For $t = 1,\ldots, \tau$, the model's memory is filled in such a manner, using the inputs $(a_t, x_t)$ (\textit{memorization} phase), so that for $t > \tau$, when only provided with $a_t$, the model can once again regress the state $s_t$ and use it to retrieve memorized features in order to predict the observation $x_t$ (\textit{anamnesis} \ie, \textit{recall} phase).

Progress has also been made towards solving visual SLAM with neural networks. %Different from SfM (Structure-from-Motion) methods~\cite{sturm1996factorization,nister2005preemptive,wu2013towards} that typically operate offline and on unordered images, classic SLAM methods~\cite{montemerlo2002fastslam,durrant2006simultaneous} usually rely on real-time continuous cues in videos and coherence among frames, and are mostly composed of several carefully engineered modules, such as tracking, mapping, and re-localization. 
CNN-SLAM~\citep{tateno2017cnnslam} replaced some modules in classical SLAM methods~\citep{durrant2006simultaneous} with neural components. Neural SLAM~\cite{zhangneural} and MapNet~\citep{henriques2018mapnet} both proposed a spatial memory system for autonomous agents. Whereas the former deeply interconnects memory operations with other predictions (\eg, motion planning), the latter offers a more generic solution with no assumption on the agents' range of action or goal. Extending MapNet, our proposed model not only attempts to build a map of the environment, but also makes incremental predictions and hallucinations based on both past experiences and current observations. %This capability of predicting under uncertainty is critical in many scenarios.
    
%\cBen{shouldn't we list works on memory networks (\cf current 1st paragraph of Methodology) \eg DND (nice solution but requires to know the agent's action $a_t$), MapNet (basis of our work - they however only cover the "write" function for the memory while we extend to "read" + extrapolate; opt. we also extend their method to 3D map), LSTM more broadly, \etc?}

%\cBen{note: we could mention LSTM Auto-encoder~\cite{zhao2018robust}}

%\cBen{GTM-SM, MapNet, AVD should still be mentioned in Related Work}

% 02
\paragraph{3D Modeling and Geometry-based View Synthesis.} Much effort has also been expended in explicitly modeling the underlying 3D structure of scenes and objects, \eg, 
%\cite{forsyth2003modern,montemerlo2002fastslam,durrant2006simultaneous,sturm1996factorization}
\citep{durrant2006simultaneous,choudhary2015information}.
While appealing and accurate results are guaranteed when multiple source images are provided, this line of work is fundamentally not able to deal with sparse inputs. 
To address this issue, \citet{flynn2016deepstereo} proposed a deep learning approach focused on the multi-view stereo problem by regressing directly to output pixel values.
On the other hand, \citet{ji2017morphing} explicitly utilized learned dense correspondences to predict the image in the middle view of two source images.
Generally, these methods are limited to synthesizing a middle view among fixed source images, whereas our framework is able to generate arbitrary target views by extrapolating from prior domain knowledge.%learn from source images vary in length.

% \paragraph{3D Voxel/Mesh/Point Cloud Prediction.} Explicitly reconstructing 3D geometry has been intensively addressed in a multi-view setting, such as SfM and SLAM \cite{forsyth2003modern, sturm1996factorization, montemerlo2002fastslam, durrant2006simultaneous},  \cite{chang2015shapenet}, predicting 3D representations such as voxels, meshes, and 3D point clouds from 2D views has achieved encouraging results \cite{choy20163d, fan2017point, lin2018learning}.
% By contrast, we are interested in synthesizing views instead of 3D representations of objects. Our approach requires no 3D supervision nor explicit 3D model.

\paragraph{Novel View Synthesis.} 
The problem we tackle here can be formulated as a novel view synthesis task: 
given pictures taken from certain poses, solutions need to synthesize an image from a new pose, and has seen significant interest in both vision \citep{park2017transformation,zhou2016view} and graphics
% \cite{hedman2016scalable,he1998layered}
\cite{hedman2016scalable}. There are two main flavors of novel view synthesis methods. The first type synthesizes pixels from an input image and a pose change with an encoder-decoder structure~\citep{tatarchenko2016multi}.
The second type reuses pixels from an input image with a sampling mechanism. For instance, \citet{zhou2016view} recasted the task of novel view synthesis as predicting dense flow fields that map the pixels in the source view to the target view, but their method is not able to hallucinate pixels missing from source view.
%Park \etal~\cite{park2017transformation} predicted a flow to move the pixels from the source to the target view, followed by an image completion network.
Recently, methods that use geometry information have gained popularity, as they are more robust to large view changes and resulting occlusions~\citep{park2017transformation}.
%\cite{yang2015weakly, tatarchenko2016multi} propose to directly generate pixels of a target view, while \cite{zhou2016view} recasts the task of novel view synthesis as predicting dense flow fields that map the pixels in the source view to the target view, but it is not able to hallucinate the pixels which are missing from source view.
%\cite{park2017transformation} predicts a flow to move the pixels from the source to the target view, followed by an image completion network.
% 03
%\cite{dosovitskiy2015learning} has proposed a supervised, conditional generative model trained to generate images of chairs, tables, and cars with specified attributes which are controlled by transformation and view parameters passed to the network.
%The range of objects which can be synthesized using the framework is strictly limited to the pre-defined models used for training; the network can generate different views of these models, but cannot generalize to unseen objects to perform inference tasks
However, these conditional generative models rely on additional data to perform their target tasks. In contrast, our proposed model enables the agent to predict its own pose and synthesize novel views in an end-to-end fashion.
% Other works have introduced a clamping strategy to enforce a specific organizational structure in the latent space \cite{}.

% %-------------------------------------------------------------------------
\section{Methodology}
\label{sec:mth}
% !TeX spellcheck = en_US

% \begin{figure}[t]
% \centering
% \includegraphics[width=1\linewidth]{registration}
% \caption{\textbf{Localization and memorization}, based on%~\cite{henriques2018mapnet}
% \ MapNet.}
% \label{fig:registration}  
% \end{figure}

%\cBen{todo: mention somewhere more clearly the global constraints we are tackling}
While the current state of the art in scene registration yields satisfying results, there are several assumptions, including prior knowledge of the agent's range of actions, as well as the actions $a_t$ themselves at each time step. In this paper, we consider unknown agents, with only their observations $x_t$ provided during the memorization phase. In the spirit of the MapNet solution~\citep{henriques2018mapnet}, we use an allocentric spatial memory map. Projected features from the input observations are registered together in a coordinate system relative to the first inputs, allowing to regress the position and orientation (\ie, \textit{pose}) of the agent in this coordinate system at each step. Moreover, given viewpoints and camera intrinsic parameters, features can be extracted from the spatial memory (\textit{frustum culling}) to recover views. Crucially, at each step, memory ``holes'' can be temporarily filled by a network trained to generate domain-relevant features while ensuring global consistency.
Put together (\cf Figure~\ref{fig:pipeline}), our pipeline (trainable both separately and end-to-end) can be seen as an explicit topographic memory system with localization, registration, and retrieval properties, as well as consistent memory-extrapolation from prior knowledge.
We present details of our proposed approach in this section.

\subsection{Localization and Memorization}

% \begin{figure*}[t]
% \centering
% \includegraphics[width=0.8\linewidth]{decoder_training_avd}
% \caption{\textbf{Training of the memorization and anamnesis modules}. This part of the pipeline can be seen as a particular auto-encoder (\ie, with memory), and can be trained end-to-end as such. $\mathcal{L}_{loc}$ measures the accuracy of the predicted allocentric poses \ie, training the encoding CNN to extract meaningful features and the LSTM to update the memory properly. $\mathcal{L}_{anam}$ measures the quality of the recalled views---rendered from $m_t$ using the ground-truth poses---compared to the original ones.}
% \label{fig:decoder_training}  
% \end{figure*}

\begin{figure}[t]
\centering
\includegraphics[width=1\linewidth]{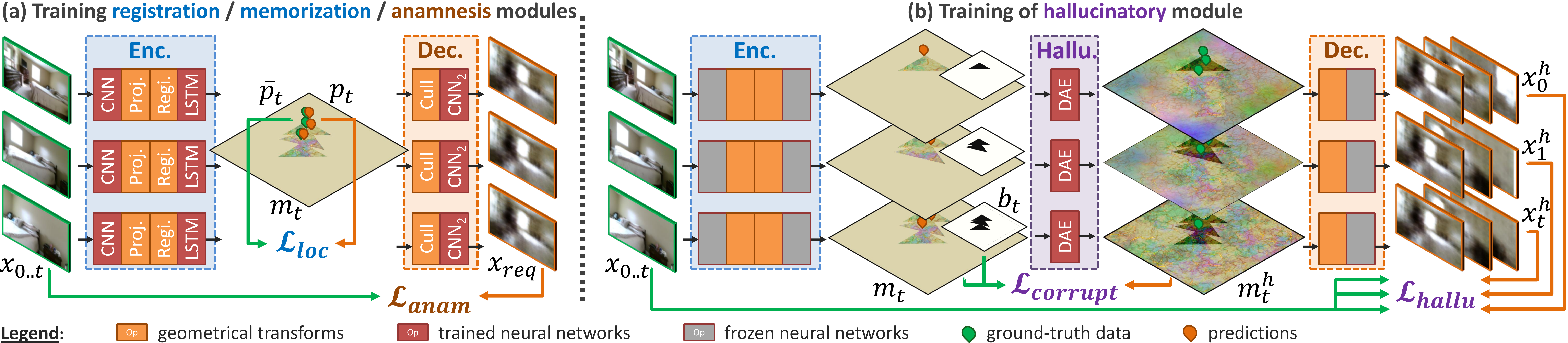}
    \vspace{-1.5em}
\caption{\textbf{Pipeline training}. Though steps are shown separately in the figure (for clarity), the method is trained in a single pass. $\mathcal{L}_{loc}$ measures the accuracy of the predicted allocentric poses, \ie, training the encoding system to extract meaningful features (CNN) and to update the global map $m_t$ properly (LSTM). $\mathcal{L}_{anam}$ measures the quality of the images rendered from $m_t$ using the ground-truth poses, to train the decoding CNN. $\mathcal{L}_{hallu}$ trains the method to predict all past and future observations at each step of the sequence, while $\mathcal{L}_{corrupt}$ punishes it for any memory corruption during hallucination.}
\label{fig:training}  
\end{figure}

Our solution first takes a sequence of observed images $x_t \in \mathbb{R}^{c \times h \times w}$ (\eg, with $c = 3$ for RGB images or $4$ for RGB-D ones) for $t = 1,\ldots, \tau$ as input, localizing them and updating the spatial memory $m \in \mathbb{R}^{n \times u \times v}$ accordingly. 
The memory $m$ is a discrete global map of dimensions $u \times v$ and feature size $n$. $m_t$ represents its state at time $t$, after updating $m_{t-1}$ with features from $x_t$.
\paragraph{Encoding Memories.}
~\label{sec:encoding}
Observations are encoded to fit the memory format. For each observation, a feature map $x'_t \in \mathbb{R}^{n \times h' \times w'}$ is extracted by an encoding convolutional neural network (CNN). % (with $n$ the feature size).
Each feature map is then projected from the 2D image domain into a tensor $o_t \in \mathbb{R}^{n \times s \times s}$ representing the agent's spatial neighborhood (to simplify later equations, we assume $u, v, s$ are odd). This operation is data and use-case dependent. For instance, for RGB-D observations of 3D scenes (or RGB images extended by some monocular depth estimation method, \eg,
%~\cite{eigen2015predicting,roy2016monocular,laina2016deeper,cao2017exploiting,kendall2017multi,xu2018pad}
\cite{xu2018pad}), the feature maps are first converted into point clouds using the depth values and the camera intrinsic parameters (assuming like \citet{henriques2018mapnet} that the ground plane is approximately known). They are then projected into $o_t$ through discretization and max-pooling (to handle many-to-one feature aggregation, \ie, when multiple features are projected into the same cell~\citep{qi2017pointnet}).
For 2D scenes (\ie, agents walking on an image plane), $o_t$ can be directly obtained from $x_t$ (with optional cropping/scaling).

\paragraph{Localizing and Storing Memories.}
\label{sec:mem-update}
Given a projected feature map $o_t$ and the current memory state $m_{t-1}$, the registration process involves densely matching $o_t$ with $m_{t-1}$, considering all possible positions and rotations.
As explained in~\citet{henriques2018mapnet}, this can be efficiently done through cross-correlation. % (\cf Figure~\ref{fig:registration}). 
Considering a set of $r$ yaw rotations, a bank $o'_t \in \mathbb{R}^{r \times n \times s \times s}$ is built by rotating $o_{t}$ $r$ times: $o'_t = \big\{R(o_t\ ,\ 2\pi \frac{i}{r}\ , \ c_{s,s})\big\}_{i=0}^r$,
%\begin{equation}
%o'_t = \big\{R(o_t\ ,\ 2\pi \frac{i}{r}\ , \ c_{s,s})\big\}_{i=0}^r
%\end{equation}
with $c_{s,s} = (\frac{s+1}{2}, \frac{s+1}{2})$ horizontal center of $o_t$, and $R(o, \alpha, c)$ the function rotating each element in $o$ around the position $c$ by an angle $\alpha$, in the horizontal plane.
% (if \eg, $r$ roll and $r$ pitch rotations are considered, the stack $o'_t$ could be extended accordingly with similar operations, into a tensor of shape $r^3 \times n \times z \times s \times s$).
% \begin{figure*}[t]
% \centering
% \includegraphics[width=0.8\linewidth]{hallu_training_avd}
% \caption{\textbf{Training of the hallucination module}. The goal of this sub-network is to fill the ``holes'' in the global memory/map, hallucinating domain-relevant features which could be used to generate new views (while keeping existing/memorized features uncorrupted). Therefore, it is trained to predict all future observations at each step ($\mathcal{L}_{hallu}$), while being punished for any corruption to the global map and recalled observations ($\mathcal{L}_{corrupt}$).}
% \label{fig:hallu_training}  
% \end{figure*}
The dense matching can therefore be achieved by sliding this bank of $r$ feature maps across the global memory $m_{t-1}$ and comparing the correlation responses. 
The localization probability field $p_t \in \mathbb{R}^{r \times u \times v}$ is efficiently obtained by computing the cross-correlation (\ie, ``\textit{convolution}", operator $\star$, in deep learning literature) between $m_{t-1}$ and $o'_t$ and normalizing the response map (\textit{softmax} activation $\sigma$).
The higher a value in $p_t$, the stronger the belief the observation comes from the corresponding pose. Given this probability map, it is possible to register $o_t$ into the global map space (\ie, rotating and translating it according to $p_t$ estimation) by directly convolving $o_t$ with $p_t$. This registered feature tensor $\hat{o}_t \in \mathbb{R}^{n \times u \times v}$ can finally be inserted into memory:
%\begin{align}
%\hat{o}_t   &= p_t \ast o'_t \quad \text{with}\;\; p_t = \sigma(m_{t-1} \star o'_t)\\
%m_t         &= \lstm(m_{t-1},\ \hat{o}_t,\ \theta_{lstm})
%\end{align}
\begin{equation}
m_t = \lstm(m_{t-1},\ \hat{o}_t,\ \theta_{lstm})
 \quad \text{with}\;\; \hat{o}_t   = p_t \ast o'_t
 \;\; \text{and}\;\;  p_t = \sigma(m_{t-1} \star o'_t)
\end{equation}
A long short-term memory (LSTM) unit is used, to update $m_{t-1}$ (the unit's \textit{hidden} state) with $\hat{o}_t$ (the unit's input) in a knowledgeable manner (\cf trainable parameters $\theta_{lstm}$). During training, the recurrent network will indeed learn to properly blend overlapping features, and to use $\hat{o}_t$ to solve potential uncertainties in previous insertions (uncertainties in $p$ result in blurred $\hat{o}$ after convolution). The LSTM is also trained to update an occupancy mask of the global memory, later used for constrained hallucination (\cf Section~\ref{sec:hallu}).

\paragraph{Training.}
The aforementioned process is trained in a supervised manner given the ground-truth agent's poses. For each sequence, the feature vector $o_{t=0}$ from the first observation is registered at the center of the global map without rotation (origin of the allocentric system). Given $\bar{p}_t$, the one-hot encoding of the actual state at time $t$, the network's loss $\mathcal{L}_{loc}$ at time $\tau$ is computed over the remaining predicted poses using binary cross-entropy:
\begin{equation}
\mathcal{L}_{loc} = -\frac{1}{\tau}\displaystyle\sum_{t=1}^{\tau}{\big[\bar{p_t} \cdot \log(p_t) + (1-\bar{p}_t) \cdot \log(1-p_t)\big]}
\end{equation}
%\cBen{small doubt here - double check}

\subsection{Anamnesis}
\label{sec:anam}

Applying a novel combination of geometrical transforms and decoding operations, memorized content can be recalled from $m_{t}$ and new images from unexplored locations synthesized.
This process can be seen as a many-to-one recurrent generative network, with image synthesis conditioned on the global memory and the requested viewpoint. We present how the entire network can thus be advantageously trained as an auto-encoder with a recurrent neural encoder and a persistent latent space.

\paragraph{Culling Memories.}
\label{sec:culling}
%\begin{figure}[t]
%\centering
%\includegraphics[width=0.5\linewidth]{culling_geometry}
%\caption{\textbf{Geometrical Memory Culling} (2D representation). The feature vector $o_{req}$, defining the observation for an agent positioned at $l_{req}$ and rotated by an angle $\alpha_{req}$ in the allocentric system, is extracted from $m_t$ through a series of geometrical transforms (rotation, translation, clipping, culling). Coordinates and distances in the allocentric coordinate system are represented in white; in yellow for the $m_t$ matrix system; and red for the $o_t$ one.}
%\label{fig:culling_geometry}  
%\end{figure}
While a decoder can retrieve observations conditioned on the full memory and requested pose, it would have to disentangle the visual and spatial information itself, which is not trivial to learn (\cf ablation study in Section~\ref{sec:exp_2d}). Instead, we propose to use the spatial properties of our memory to first  \textit{cull} features from requested viewing volumes before passing them as inputs to our decoder. More formally, given the allocentric coordinates $l_{req} = (u_{req}, v_{req})$%
%\footnote{For concision, the demonstration is done on a 2D map \ie, $z=1$. Extension to 3D data is explained in the supplementary material.}
, orientation $\alpha_{req} = 2\pi\frac{r_{req}}{r}$, and field of view $\alpha_{fov}$, $o_{req} \in \mathbb{R}^{n \times s \times s}$ representing the requested neighborhood is filled as follow:
\begin{equation}
o_{req,kij} = \begin{cases}
                   \hat{o}_{req,kij} \quad &\text{if } \atantwo{\frac{j - \frac{s+1}{2}}{i - \frac{s+1}{2}}} < \frac{\alpha_{fov}}{2} \\
                   -1 \quad &\text{otherwise}
                 \end{cases}\\
\end{equation}
with $\hat{o}_{req}$ the unculled feature patch extracted from $m_t$ rotated by $-\alpha_{req}$, \ie, $\forall k \in [0 \twodots n-1],\ \forall (i,j) \in [0 \twodots s-1]^2$:
%\begin{aligned}
%\hat{o}_{req, kij} &= R(m_t,\ -\alpha_{req},\ c_{u,v} + l_{req})_{k \xi \eta}\\
%(\xi, \eta) &= (i,j) +  c_{u,v} + l_{req} - c_{s,s}
%\end{aligned}
\begin{equation}
\hat{o}_{req, kij} = R(m_t,\ -\alpha_{req},\ c_{u,v} + l_{req})_{k \xi \eta}
\quad \text{with}\quad (\xi, \eta) = (i,j) +  c_{u,v} + l_{req} - c_{s,s}
\end{equation}
%\cBen{fix/improve equations}

This differentiable operation combines feature extraction (through translation and rotation) and \textit{viewing frustum culling} (\cf computer graphics to render large 3D scenes).

\begin{figure}[t]
\centering
\includegraphics[width=\linewidth]{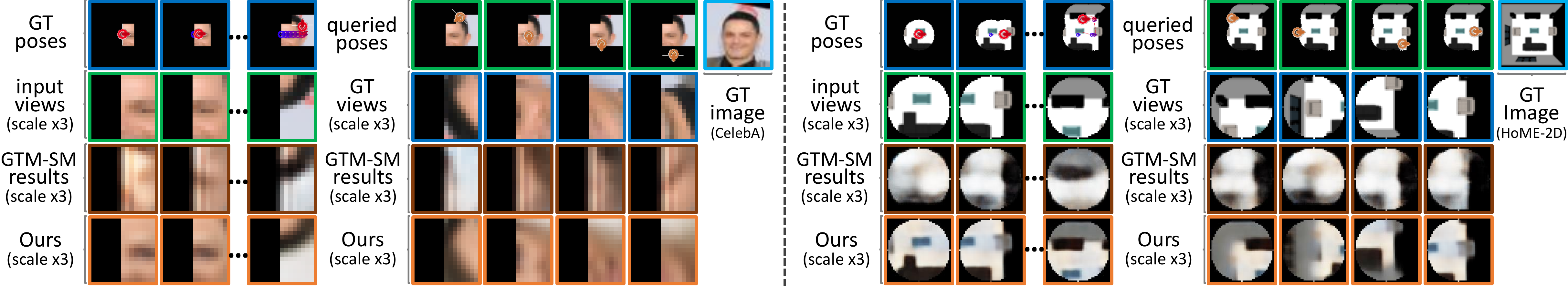}
    \vspace{-1.5em}
\caption{\textbf{Synthesis of memorized and novel views%~\citep{fraccaro2018generative} 
}%~\citep{liu2015faceattributes}
\ from 2D scenes, comparing to GTM-SM~\citep{fraccaro2018generative}. 
% UNCOMMENTING THE LONGER VERSION OF THIS LEGEND:
Methods receive a sequence of 10 observations (along with the related actions for GTM-SM) from an exploring agent, then they apply their knowledge to generate 46 novel views. GTM-SM has difficulties grasping the structure of the environment from short observation sequences, while our method usually succeeds thanks to prior knowledge.
% Our method benefits from prior knowledge and global representation.
}
\label{fig:celeba_gtmsm_comp2}  
\end{figure}

\paragraph{Decoding Memories.}
As input observations undergo encoding and projection, feature maps culled from the memory go through a reverse procedure to be projected back into the image domain. With the synthesis conditioning covered in the previous step, a decoder directly takes $o_{req}$ (\ie, the view-encoding features) and returns $x_{req}$, the corresponding image.
This back-projection is still a complex task. The decoder must both project the features from voxel domain to image plane, and decode them into visual stimuli. Previous works
%(\eg, 
%~\citep{qi2017pointnet,kato2018neural}
%~\citep{kato2018neural}
%)
and qualitative results demonstrate that a well-defined (\eg, \textit{geometry-aware}) network can successfully accomplish this task.

\paragraph{Training.}
By requesting the pipeline to recall given observations---\ie, setting $l_{req, t} = \bar{l}_t$ and $r_{req, t} = \bar{r}_t$, $\forall t \in [1, \tau]$, with $\bar{l}_t$ and $\bar{r}_t$ the agent's ground-truth position/orientation at each step $t$---it can be trained end-to-end as an image-sequence auto-encoder (\cf Figure
%~\ref{fig:decoder_training})
\ref{fig:training}.a). 
% by means of the following loss:
% \begin{equation}
% \mathcal{L}_{anam} = \frac{1}{\tau}\displaystyle\sum_{t=0}^{\tau}{|x_t - x_{req, t}|_1}
% \end{equation}
%\cBen{I've replaced the L1 loss equation by text... It's not as clear, but saves lots of space...?}
Therefore, its loss $\mathcal{L}_{anam}$ is computed as the L1 distance between $x_t$ and $x_{req, t}$, $\forall t \in [0, \tau]$, averaged over the sequences.
% Taking advantage of our framework's modularity, the encoder and decoder can even be pre-trained together, \ie, removing the global map and registration process in between, passing instead the feature maps directly from one sub-network to another as a traditional auto-encoder. We observe that such a pre-training tends to stabilize the overall learning process, given the correlation between features for auto-encoding and features for registration.
Note that thanks to our framework's modularity, the global map and registration steps can be removed to pre-train the encoder and decoder together (passing the features directly from one to the other). We observe that such a pre-training tends to stabilize the overall learning process.

\subsection{Mnemonic Hallucination}
\label{sec:hallu}

While the presented pipeline can generate novel views% (\ie, from poses not explored by the agent yet)
, these views have to overlap with previous observations for the solution to extract enough features for anamnesis.
Therefore, we extend our memory system with an \textit{extrapolation} module%. Given prior domain-relevant knowledge (\ie, an estimation of the environments distribution learned during training), this module can
\ to \textit{hallucinate} relevant features %to fill regions not yet covered by the agent.
for unexplored regions.

\paragraph{Hole Filling with Global Constraints.}
\label{sec:hole-fill}
Under global constraints, we build a deep auto-encoder (DAE) in the feature domain, which takes $m_t$ as input, as well as a noise vector of variable amplitude (\eg, no noise for deterministic navigation planning or heavy noise for image dataset augmentation), and returns a convincingly hole-filled version $m^h_t$, while leaving registered features uncorrupted. In other words, this module should provide relevant features while seamlessly integrating existing content according to prior domain knowledge.
% In order to improve the sampling of hallucinated features and the global awareness of this generator, we adopt several concepts from SAGAN~\citep{zhang2018self}. Following the generative adversarial network (GAN) strategy~\citep{goodfellow2014generative,radford2015unsupervised,salimans2016improved,isola2017pix2pix}, our conditioned generator is trained against a discriminator evaluating the \textit{realism} of feature patches $o^h_t$ culled from $m^h_t$. This discriminator is itself trained against $o_t$ (\textit{real} samples) and $o^h_t$ (\textit{fake} ones). Furthermore, both generator and discriminator contain self-attentive layers~\citep{zhang2018self,cheng2016long,parikh2016decomposable,vaswani2017attention}, which can efficiently model relationships between widely-separated spatial regions. 
%Given a feature map $o \in \mathbb{R}^{n \times u \times v}$, the result $o_{sa}$ of the self-attention operation is:
%\begin{equation}
%o_{sa} = o + \gamma \sigma\big((W_f \star o)^\intercal (W_g \star o)\big) %\cdot (W_h \star o)
%\end{equation}
%with $W_f \in \mathbb{R}^{\bar{n} \times n}$, $W_g \in \mathbb{R}^{\bar{n} \times n}$, $W_h \in \mathbb{n}^{n \times n}$ learned weight matrices (we opt for $\bar{n} = \sfrac{n}{8}$ as in~\citep{zhang2018self}); and $\gamma$ a trainable scalar weight.

\paragraph{Training.}
Assuming the agent homogeneously explores training environments, the hallucinatory module is trained at each step $t \in [0, \tau_{cur}]$ by generating $m^h_t$, the hole-filled memory used to predict yet-to-be-observed views $\{x_t\}_{t=\tau_{cur}+1}^\tau$. To ensure that registered features are not corrupted, we also verify that all observations $\{x_t\}_{t=0}^{\tau_{cur}}$ can be retrieved from $m^h_t$ (\cf Figure
%~\ref{fig:hallu_training})
\ref{fig:training}.b). 
This generative loss is computed as follows:
\begin{equation}
\mathcal{L}_{hallu} = \frac{1}{\tau(\tau - 1)}\displaystyle\sum_{t=0}^{\tau-1}{\displaystyle\sum_{i=0}^{\tau}{|x^h_i - x_i|_1}}
\end{equation}
with $x^h_i$ the view recovered from $m^h_i$ using the agent's true location $\bar{l}_i$ and orientation $\bar{r}_i$ for its observation $x_i$.
Additionally, another loss is directly computed in the feature domain, using memory occupancy masks $b_t$ to penalize any changes to the registered features (given $\odot$ Hadamard product):
\begin{equation}
\mathcal{L}_{corrupt} = \frac{1}{\tau}\displaystyle\sum_{t=0}^{\tau}{\lvert(m^h_t - m_t) \odot b_t\rvert_1}
\end{equation}
Trainable end-to-end, our model efficiently acquires domain knowledge to register, hallucinate, and synthesize scenes.
% These losses are used to optimize the hallucinatory module only, and are not back-propagated to the rest of the pipeline. The complete system can be trained with a single forward pass (registration, hallucination, anamnesis) and two back-propagation steps.

% %-------------------------------------------------------------------------
\section{Experiments}
\label{sec:exp}
% \input{content-experiments-for-preprint}
% !TeX spellcheck = en_US

% As described in Section~\ref{sec:mth}, 
We demonstrate our solution on various synthetic and real 2D and 3D environments%~\footnote{Implementation details and further quantitative results are in the supplementary material, accessible here: \url{http://bit.do/iss-supmat}.}
. For each experiment, we consider an unknown agent exploring an environment, only providing a short sequence of partial observations (limited field of view). Our method has to localize and register the observations, and build a global representation of the scene. Given a set of requested viewpoints, it should then render the corresponding views. In this section, we qualitatively and quantitatively evaluate the predicted trajectories and views, comparing with GTM-SM~\citep{fraccaro2018generative}, the only other end-to-end %(and only reproducible)
memory system for scene synthesis, based on the Generative Query Network~\citep{eslami2018neural}.
%\cBen{not sure the last part of this paragraph is a good idea...?}

%\subsection{Implementation Details}

%\cBen{Write proerply and concise. All networks are ResNet4. Self-attention layers and spectal norm used for hallu GAN c.f. SAGAN. 16 channels for global maps. More details in sup-mat}

\subsection{Navigation in 2D Images}
\label{sec:exp_2d}

We first study agents exploring images (randomly walking, accelerating, rotating), observing the image patch in their field of view at each step (more details and results in the supplementary material).

\paragraph{Experimental Setup.}
%\begin{figure}[t]
%\centering
%\includegraphics[width=\linewidth]{figures/celeba_gtmsm_comp.pdf}
%\caption{\textbf{Qualitative comparison with %GTM-SM~\citep{fraccaro2018generative} on CelebA %dataset}~\citep{liu2015faceattributes}. GTM-SM has difficulties grasping %the structure of the environment from short observation sequences (10 %observations here), while our method usually succeeds thanks to prior %knowledge.}
%\label{fig:celeba_gtmsm_comp}  
%\end{figure}
We use a synthetic dataset of indoor $83 \times 83$ floor plans rendered using the HoME platform~\citep{brodeur2017home}  and SUNCG data~\citep{SUNCG} (8,640 training + 2,240 test images from random rooms ``\textit{office}", ``\textit{living}", and ``\textit{bedroom}"). Similar to~\citet{fraccaro2018generative}, we also consider an agent exploring real pictures from the CelebA dataset~\citep{liu2015faceattributes}, scaled to $43 \times 43$px. We consider two types of agents for each dataset. %,  with more or less realistic characteristics. 
To reproduce \citet{fraccaro2018generative} experiments, we first consider non-rotating agents $A^s$---only able to translate in the 4 directions---with a $360^\circ$ field of view covering an image patch centered on the agents' position. The CelebA agent $A^s_{cel}$ has a $15 \times 15$px square field of view; while the field of view of the HoME-2D agent $A^s_{hom}$ reaches $20$px away, and is therefore circular (in the $41 \times 41$ patches, pixels further than $20$px are left blank).
To consider more complex scenarios, agents $A^c_{cel}$ and $A^c_{hom}$ are also designed. They can rotate and translate (in the gaze direction), observing patches rotated accordingly. On CelebA images, $A^c_{cel}$ can rotate by $\pm 45^\circ$ or $\pm 90^\circ$ each step, and only observes $8 \times 15$ patches in front ($180^\circ$ rectangular field of view); while for HoME-2D, $A^c_{hom}$ can rotate by $\pm 90^\circ$ and has a $150^\circ$ field of view limited to $20$px.
All agents can move from $\sfrac{1}{4}$ to $\sfrac{3}{4}$ of their field of view each step. Input sequences are 10 steps long. For quantitative studies, methods have to render views covering the whole scenes \wrt the agents' properties.

\begin{figure}[t]
\centering
\includegraphics[width=\linewidth]{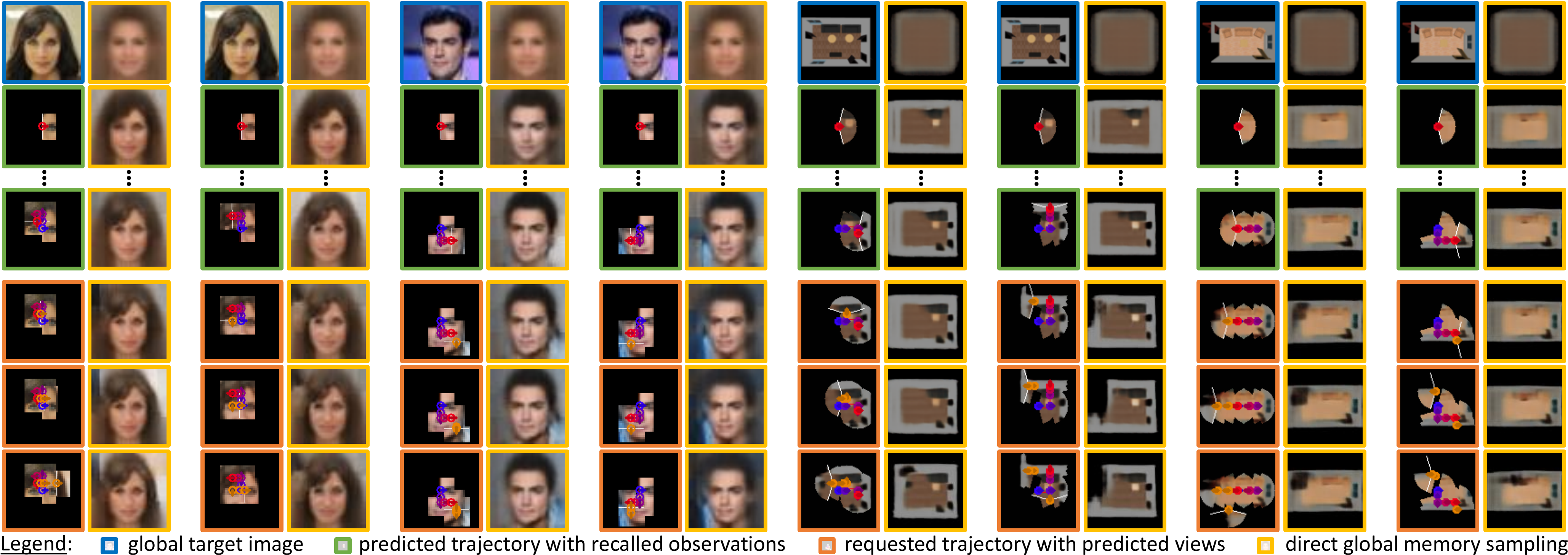}
\vspace{-1.5em}
\caption{\textbf{Incremental exploration and hallucination} 
%(on CelebA/HoME-2D)%~\citep{brodeur2017home}
(on 2D data). Scene representations evolve with the registration of observed or hallucinated views  (\eg, adapting hair color, face orientation,\etc).}
\label{fig:celebahome_incremental_hallu}  
\end{figure}

\begin{figure}[t]
\centering
\includegraphics[width=1\linewidth]{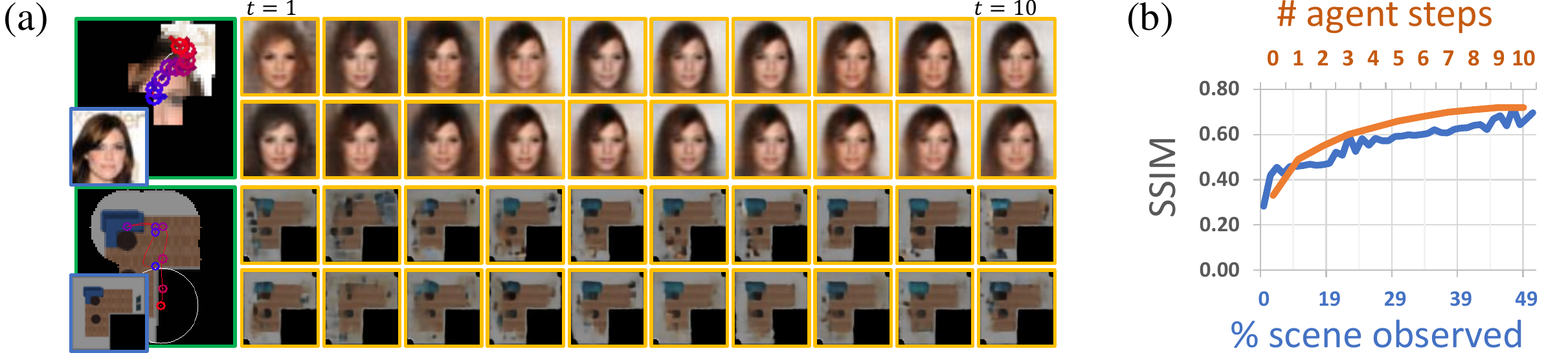}
%\vspace{-1.8em}
\vspace{-1.5em}
\caption{
%\textbf{Stochastic Hallucinations.} 
(a) \textbf{Statistical nature of the hallucinated content}. Global scene representations are shown for each step $t$, given the same agent trajectories but different noise vectors passed to the hallucinatory auto-encoder; (b) \textbf{Salient image quality \wrt agent steps and scene coverage} for $A^s_{cel}$, computed over the global scene representations. These results show how the global scene properties converge and the quality of generated images increase as observations accumulate.}
\label{fig:rebuttal_noise}  
\end{figure}

\paragraph{Qualitative Results.}
As shown in Figure~\ref{fig:celeba_gtmsm_comp2}, our method efficiently uses prior knowledge to register observations and extrapolate new views, consistent with the global scene and requested viewpoints. While an encoding of the agent's actions is also provided to GTM-SM%~\citep{fraccaro2018generative}
\ (guiding the localization), it cannot properly build a global representation from short input sequences, and thus fails at rendering completely novel views. Moreover, unlike the dictionary-like memory structure of GTM-SM, our method stores its representation into a single feature map, which can therefore be queried in several ways. As shown in Figure~\ref{fig:celebahome_incremental_hallu}, for a varying number of conditioning inputs, one can request novel views one by one, culling and decoding features; with the option to register hallucinated views back into memory  (\ie, saving them as ``valid" observations to be reused). But one can also directly query the full memory, training another decoder to convert all the features. Figure~\ref{fig:celebahome_incremental_hallu} also demonstrates how different trajectories may lead to different intermediate representations, while Figure~\ref{fig:rebuttal_noise}-a illustrates how the proposed model can predict different global properties for identical trajectories but different hallucinatory noise.
In both cases though (different trajectories or different noise), the scene representations converge as the scene coverage increases.

\begin{table*}[t]
	\small
	\centering
	\begin{ThreePartTable}
	\caption{
		\textbf{Quantitative comparison on 2D and 3D scenes}, \cf setups in Subsections~\ref{sec:exp_2d}-\ref{sec:3D_exp} \textit{($\searrow$ the lower the better; $\nearrow$ the higher the better; ``$u$" horizontal bin unit according to AVD setup)}.
	}
    \vspace{0.1em}
	\label{tab:quanti_celebA_complex} 
	\resizebox{1\linewidth}{!}{
		\def\arraystretch{1}
		%\footnotesize
		\begin{tabular}{@{}l|c|ccc|ccc|cc|cc@{}}
			\toprule
	        
			  \multicolumn{1}{c|}{\multirow{2}{*}{\textbf{Exp.}}}
			& \multirow{2}{*}{\textbf{Methods}}
			& \multicolumn{3}{c|}{\textbf{Average Position Error}}
			& \multicolumn{3}{c|}{\textbf{Absolute Trajectory Error}}
			& \multicolumn{2}{c|}{\textbf{Anam. Metr.}}
			&  \multicolumn{2}{c}{\textbf{Hall. Metr.}}
			\\ \cline{3-12}
			& 
			& \rule[5pt]{0pt}{5pt} \textbf{Med.$^\searrow$} & \textbf{Mean$^\searrow$}  & \textbf{Std.$^\searrow$} 
			& \rule[5pt]{0pt}{5pt} \textbf{Med.$^\searrow$} & \textbf{Mean$^\searrow$} & \textbf{Std.$^\searrow$} 
			& \rule[5pt]{0pt}{5pt} \textbf{L1$^\searrow$} & \textbf{SSIM$^\nearrow$}
			& \rule[5pt]{0pt}{5pt} \textbf{L1$^\searrow$} & \textbf{SSIM$^\nearrow$}
			
			\\
			\midrule 
			%\multirow{2}{*}{\textbf{A) CelebA, \boldmath$A^s_{cel}$}}
			\multirow{2}{*}{\textbf{A) \boldmath$A^s_{cel}$}}
			& GTM-SM & 4.0px &  4.78px & 4.32px & 6.40px & 6.86px & 3.55px & 0.14 & 0.57 & 0.14 & 0.41 \\
			& GTM-SM$^{L1_{s_t \leftrightarrow l_t}}$ \tnote{*} 
			& \textbf{1.0px} &  1.03px & 1.23px 
			& 0.79px & 0.87px & 0.86px 
			& 0.13 & 0.64 & 0.15 & 0.40 \\
			& GTM-SM$^{s_t \leftarrow l_t}$ \tnote{**}
			& \multicolumn{3}{c|}{0px (NA -- poses passed as inputs)} 
			& \multicolumn{3}{c|}{0px (NA -- poses passed as inputs)} 
			& 0.08 & 0.76 & 0.13 & 0.43 \\
			& Ours & \textbf{1.0px} & \textbf{0.68px} & \textbf{1.02px} & \textbf{0.49px} & \textbf{0.60px} & \textbf{0.64px} & \textbf{0.06} & \textbf{0.80} & \textbf{0.09} & \textbf{0.72} \\
			\midrule
			
			% \multirow{2}{*}{\textbf{B) CelebA, \boldmath$A^c_{cel}$}}
			\multirow{2}{*}{\textbf{B) \boldmath$A^c_{cel}$}}
			& GTM-SM & 3.60px & 5.04px & 4.42px & 2.74px & 1.97px & 2.48px & 0.21 & 0.50 & 0.32 & 0.41 \\
			& Ours & \textbf{1.0px} & \textbf{2.21px} & \textbf{3.76px} & \textbf{1.44px} & \textbf{1.72px} & \textbf{2.25px} & \textbf{0.08} & \textbf{0.79} & \textbf{0.20} & \textbf{0.70} \\
			\midrule 
			
			%\multirow{2}{*}{\textbf{C) HoME-2D, \boldmath$A^s_{cel}$}}
			\multirow{2}{*}{\textbf{C) \boldmath$A^s_{hom}$}}
			& GTM-SM & 4.0px &  4.78px & 4.32px & 6.40px & 6.86px & 3.55px & 0.14 & 0.57 & 0.14 & 0.41 \\
			& Ours & \textbf{1.0px} & \textbf{0.68px} & \textbf{1.02px} & \textbf{0.49px} & \textbf{0.60px} & \textbf{0.64px} & \textbf{0.06} & \textbf{0.80} & \textbf{0.09} & \textbf{0.72} \\
			\midrule  
			
			%\multirow{2}{*}{\textbf{D) AVD 3D}}
			\multirow{2}{*}{\textbf{D) Doom}}
			& GTM-SM & {1.41u} &  {2.15u} & \textbf{{1.84u}} 
			         & \textbf{1.73u} & \textbf{1.81u} & \textbf{1.06u} 
			         & {0.09} & {0.52} & {0.13} & {0.49} \\
            & Ours& \textbf{{1.00u}} & \textbf{{1.64u}} & {2.16u} 
            & {1.75u} & {1.95u} & {1.24u}
            & {0.09} & \textbf{0.56} & \textbf{{0.11}} & \textbf{{0.54}} \\
			\midrule  
			
			%\multirow{2}{*}{\textbf{D) AVD 3D}}
			\multirow{2}{*}{\textbf{E) AVD}}
			& GTM-SM & {1.00u} &  {0.77u} & {0.69u} & {0.31u} & {0.36u} & {0.40u} & {0.37} & {0.12} & {0.43} & {0.10} \\
            & Ours& \textbf{{0.37u}} & \textbf{{0.32u}} & \textbf{{0.26u}} 
            & \textbf{{0.20u}} & \textbf{{0.21u}} & \textbf{{0.18u}}
            & \textbf{{0.22}} & \textbf{{0.31}} & \textbf{{0.25}} & \textbf{{0.23}} \\
			\bottomrule
		\end{tabular}  
	}
		\begin{tablenotes}\scriptsize
            \item [*] GTM-SM$^{L1_{s_t \leftrightarrow l_t}}$: Custom GTM-SM with a L1 localization loss computed between the predicted states $s_t$ and ground-truth poses $l_t$.
            \item [**] GTM-SM$^{s_t \leftarrow l_t}$: Custom GTM-SM with the ground-truth poses $l_t$ provided as input (no $s_t$ inference).
        \end{tablenotes}
	\end{ThreePartTable}
\end{table*}

\begin{table}[t]
 	% \vspace{-0.05cm} 
	\small
	\centering
	\caption{
		\textbf{Ablation study} on CelebA with agent $A^c_{cel}$. Removed modules are replaced by identity mappings; remaining ones are adapted to the new input shapes when necessary. LSTM, memory, and decoder are present in all instances (``Localization'' is the MapNet module).
	}
	\vspace{-0.2cm}

	\label{tab:ablation_tiny} 
	\resizebox{1\linewidth}{!}{
		\fontsize{5}{5}\selectfont
		\begin{tabu}{@{}cccc|cc|cc@{}}
			\toprule
	        
			   \multicolumn{4}{c|}{\textbf{Pipeline Modules}}
			& \multicolumn{2}{c|}{\textbf{Anamnesis Metrics}}
			& \multicolumn{2}{c}{\textbf{Hallucination Metrics}}
			\\ \midrule
		\textbf{Encoder} & \textbf{Localization}  & \textbf{Hallucinatory DAE}  & \textbf{Culling} 
			& \textbf{L1$^\searrow$} & \textbf{SSIM$^\nearrow$}
			& \textbf{L1$^\searrow$} & \textbf{SSIM$^\nearrow$}
			
			\\
			\midrule 
			$\varnothing$ &$\varnothing$   &$\varnothing$   &$\varnothing$  
			    & 0.18 & 0.62 & 0.24 & 0.59 \\
			$\checkmark$ &$\varnothing$   &$\varnothing$   &$\varnothing$  
			    & 0.17 & 0.62 & 0.24 & 0.58 \\
			$\checkmark$ & $\checkmark$ &$\varnothing$   &$\varnothing$ 
			    & 0.15 & 0.66 & 0.20 & 0.61 \\
			$\checkmark$ & $\checkmark$ & $\checkmark$ &$\varnothing$
			    & 0.15 & 0.65 & 0.19 & 0.62 \\
			$\checkmark$  &$\varnothing$  & $\checkmark$ & $\checkmark$ 
			    & 0.14 & 0.69 & 0.19 & 0.63 \\
			$\varnothing$ & $\checkmark$ & $\checkmark$ & $\checkmark$ 
		        & 0.13 & 0.71 & 0.17 & 0.66 \\
			$\checkmark$ & $\checkmark$ &$\varnothing$   & $\checkmark$ 
			    & \textbf{0.08} & \textbf{0.80} & 0.18 & 0.66 \\
			$\checkmark$ & $\checkmark$ & $\checkmark$ & $\checkmark$ 
			    & \textbf{0.08} & \textbf{0.80} & \textbf{0.15} & \textbf{0.70} \\
			\bottomrule
		\end{tabu}
		\vspace{-0.5cm}   
	}
\end{table}

\begin{figure}[t]
\centering
\includegraphics[width=\linewidth]{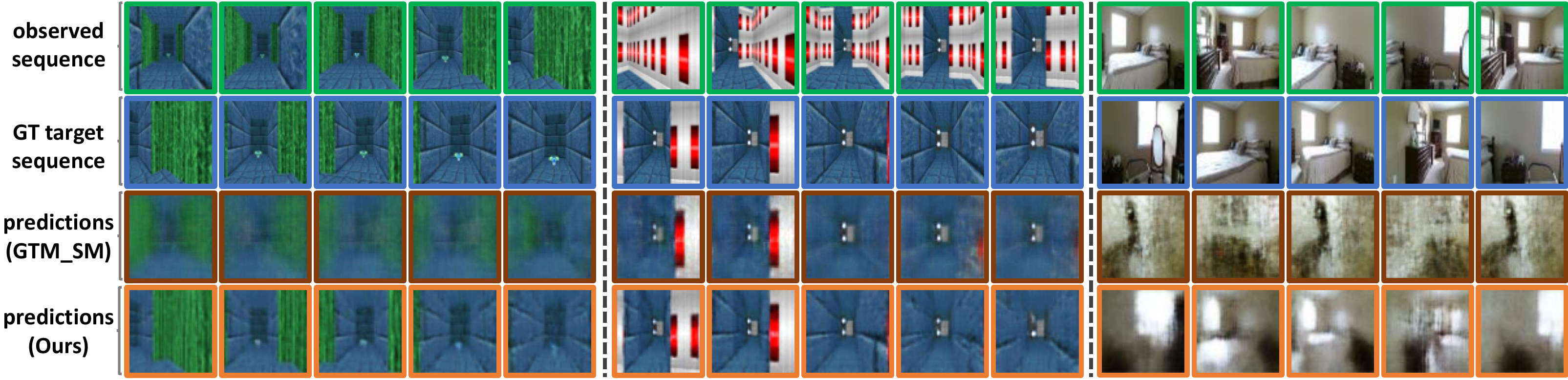}
\vspace{-1.5em}
\caption{\textbf{Qualitative comparison on 3D use-cases}, \wrt anamnesis and hallucination.
%. Methods receive sequences of 5 observations (and the corresponding actions for GTM-SM) from an agent exploring unseen test scenes, then apply their knowledge to generate novel views.
%\cf experimental setup in Section~\ref{sec:3D_exp}.
}
\label{fig:avd_results_small}  
\end{figure}

\paragraph{Quantitative Evaluations.}
We quantitatively evaluate the methods' ability to register observations at the proper positions in their respective coordinate systems (\ie, to predict agent trajectories)% \footnote{The post-processing of these relative trajectories for fair comparison is detailed in the supplementary material}
, to retrieve observations from memory, and to synthesize new ones.
% For localization, we apply two metrics commonly used to evaluate SLAM and tracking systems~\citep{henriques2018mapnet,julier2003stability,sturm2012benchmark,mur2015orb}. The average position error (APE) computes the mean Euclidean distance between the predicted positions and their ground-truths for each sequence. The absolute trajectory error (ATE) is obtained by calculating the root-mean-squared error in the positions of each sequence, after transforming the predicted trajectory to best fit the ground-truth (giving an advantage to GTM-SM predictions through post-processing, as explained in the supplementary material). 
%\cBen{I commented the definition of all metrics. Not sure if fine}
For localization, we measure the average position error (APE) and the absolute trajectory error (ATE), commonly used to evaluate SLAM systems~\citep{choudhary2015information}. 

For image synthesis, we make the distinction between recalling images already observed (\textit{anamnesis}) and generating unseen views (\textit{hallucination}). 
% For each case, two metrics are used. The common L1 distance is computed as the per-pixel absolute difference between the predicted and expected values, averaged over each image. The structural similarity ($\SSIM$) index~\citep{wang2003multiscale,wang2004image}, prevalent in the assessment of perceptual quality~\citep{wang2011information,zhang2012comprehensive,gore2015full}, is computed over $N \times N$ windows extracted from the predicted and ground-truth images, as follow:
% \begin{equation}
%   \SSIM(x,\bar{x}) = \frac{(2\mu_x\mu_{\bar{x}} + c_1) + (2 \sigma _{x\bar{x}} + c_2)} 
%     {(\mu_x^2 + \mu_{\bar{x}}^2+c_1) (\sigma_x^2 + \sigma_{\bar{x}}^2+c_2)}
% \end{equation}
% with $x$ and $\bar{x}$ the windows extracted from the predicted and ground-truth images, $\mu_x$ and $\mu_{\bar{x}}$ the mean values of the respective windows, $\sigma_x^2$ and $\sigma_{\bar{x}}^2$ their respective variance, $c_1 = 0.001^2$ and $c_2 = 0.003^2$ two constants for numerical stability. The final index is computed by averaging the values obtained by sliding the windows over the whole images (no overlapping). We opt for $N=5$ for the CelebA experiments, and $N=13$ for the HoME-2D ones (\ie, splitting the observations in 9 windows). Note that the closer to $1$ the computed index, the better the perceived image quality.
For both, we compute the common L1 distance between predicted and expected values, and the structural similarity ($\SSIM$) index~\citep{wang2003multiscale} for the assessment of perceptual quality~\citep{wang2011information,zhang2012comprehensive}.

% Tables~\ref{tab:quanti_celebA_simple}-\ref{tab:quanti_celebA_complex}-\ref{tab:quanti_home2d_simple} contain the results for the various 2D experiments. 
Table~\ref{tab:quanti_celebA_complex}.A-C shows the comparison on 2D cases. For pose estimation, our method is generally more precise even though it leverages only the observations to infer trajectories, whereas GTM-SM also infers more directly from the provided agent actions. However, GTM-SM is trained in an unsupervised manner, without any location information. Therefore, we extend our evaluation by comparing our method with two custom GTM-SM solutions that leverage ground-truth poses during training (supervised L1 loss over the predicted states/poses) and inference (poses directly provided as additional inputs). While these changes unsurprisingly improve the accuracy of GTM-SM, our method is still on a par with these results (\cf Table~\ref{tab:quanti_celebA_complex}.A).

Moreover, while GTM-SM fares well enough in recovering seen images from memory, it cannot synthesize views out of the observed domain. Our method not only extrapolates adequately from prior knowledge, but also generates views which are consistent from one to another (\cf Figure~\ref{fig:celebahome_incremental_hallu} showing views stitched into a consistent global image). Moreover, as the number of observations increases, so does the quality of the generated images (\cf Figure~\ref{fig:rebuttal_noise}-b), Note that on a Nvidia Titan X, the whole process (registering 5 views, localizing the agent, recalling the 5 images, and generating 5 new ones) takes less than 1s.

\paragraph{Ablation Study.}
Results of an ablation study are shown in Table~\ref{tab:ablation_tiny} to further demonstrate the contribution of each module. % to the image quality and global consistency. 
Note that the APE/ATE are not represented, as they stay constant as long as the MapNet localization is included. In other words, our extensions cause no regression in terms of localization. 
% Note also that without $\mathcal{L}_{corrupt}$, the anamnesis SSIM decreases by {\raise.17ex\hbox{$\scriptstyle\mathtt{\sim}$}}$0.3$.
Localizing and clipping features facilitate the decoding process by disentangling the visual and spatial information, thus improving the synthesis quality. Hallucinating features directly in the memory ensures image consistency.
%(localization and clipping guide the decoder in generating localized views, hallucination of missing features over the complete memory ensures consistency, \etc).

% \begin{figure*}[t]
% \centering
% \includegraphics[width=\linewidth]{home_incremental_hallu.pdf}
% \caption{\textbf{Incremental recovery of global representation}, illustrated with the HoME-2D experiment~\citep{brodeur2017home}. After each observation our pipeline localizes and registers, its global representation of the environment evolves accordingly (\eg, adapting the overall hair or background color, face orientation, \etc). Therefore, different trajectories lead to different intermediate representations, though they converge as the environment coverage increases.}
% \label{fig:home_incremental_hallu}  
% \end{figure*}

%\begin{figure*}[t]
%\centering
%\includegraphics[width=\linewidth]{home2d_qualitative}
%\caption{\textbf{Qualitative Results on HoME-2D dataset}~\citep{brodeur2017home}. An agent with a realistic limited view field ($150^\circ$, circular) explores 2D rooms, building a global representation and extrapolating the layout of unexplored areas.}
%\label{fig:home2d_qualitative}  
%\end{figure*}

\subsection{Exploring Virtual and Real 3D Scenes}
\label{sec:3D_exp}

We finally demonstrate the capability of our method on the more complex case of 3D scenes. 

\paragraph{Experimental Setup.}
% \cBen{Explain properly: RGB-D dataset of indoor scenes, with  pics every 30cm / 30deg ; which scenes are used for training and which for testing ; sequences obtained by letting the agent's choose between 3 (?) actions : translate 30cm closer (prob 0.7), rotate 30deg clockwise, rotate 30deg counterclockwise; RGB images normalized between -1 and 1 ; depth values converted to mm ; scaled down to $64 \times 64$, adapting the camera's intrinsics parameters accordingly for 3D projection ; sequences of 5 observations + 5 new viewpoints for NVS}
% \cRong{Working on this now...}
As a first 3D experiment, we recorded, with the Vizdoom platform~\citep{wydmuch2018vizdoom}, 34 training and 6 testing episodes of 300 RGB-D observations from a human-controlled agent navigating in various static virtual scenes (walking with variable speed or rotating by $30^\circ$ each step). Poses are discretized into 2D bins of $30 \times 30$ game units. Trajectories of 10 continuous frames are sampled and passed to the methods (the first 5 images as observations, and the last 5 as training ground-truths). We then consider the Active Vision Dataset (AVD) \citep{active-vision-dataset2017} which covers various real indoor scenes, often capturing several rooms per scene. We selected $15$ for training and $4$ for testing as suggested by the dataset authors, for a total of {\raise.17ex\hbox{$\scriptstyle\mathtt{\sim}$}}$20,000$ RGB-D images densely captured every $30$cm (on a 2D grid) and every $30^\circ$ in rotation. For each scene we randomly sampled $5,000$ agent trajectories of 10 frames each (each step the agent goes forward with $70\%$ probability or rotates either way, to favor exploration).
%When sampling trajectories, the agent goes forward with a $70\%$ probability, otherwise rotates clockwise or counterclockwise (to favor exploration). 
For both experiments, the 10-frame sequences are passed to the methods---the first 5 images as observations and the last 5 as ground-truths during training.
Again, GTM-SM also receives the action encodings. % (``\textit{rotate clockwise}'', ``\textit{rotate counterclockwise}'', ``\textit{translate forward}'', \etc).
For our method, we opted for $m \in \mathbb{R}^{32 \times 43 \times 43}$ for the Doom setup and $m \in \mathbb{R}^{32 \times 29 \times 29}$ for the AVD one.
% For each observed RGB and depth image pair, the RGB image is normalized between $[-1, 1]$, while the depth image is loaded as is and converted to the proper millimeter unit (absolute values are needed for the ground-projection processing). This data preparation mechanism follows the spirit of 2D experiments, and is consistent for both GTM-SM and our method.
\paragraph{Qualitative Results.}
Though a denser memory could be used for more refined results, Figure~\ref{fig:avd_results_small} shows that our solution is able to register meaningful features and to understand scene topographies simply from 5 partial observations. We note that quantization in our method is an application-specific design choice rather than a limitation. When compute power and memory allow, finer quantization can be used to obtain better localization accuracy (\cf comparisons and discussion presented by MapNet authors~\citep{henriques2018mapnet}). In our case, relatively coarse quantization is sufficient for scene synthesis, where the global scene representation is more crucial. In comparison, GTM-SM generally fails to adapt the VAE prior and predict the belief of target sequences (refer to the supplementary material for further results).

\paragraph{Quantitative Evaluation.}
% \cBen{Add: table of quantitative results~\ref{tab:quanti_avd} for our method and GTM-SM + opt. M2N on AVD ; reference Section~\ref{sec:sup-mat-quati2D} which contains the metrics descriptions}
% \cBen{Explain properly: It confirms the observations made on the quali results.}
%For comparison, the same metrics as in Section~\ref{sec:exp_2d} are adopted, to measure the registration and synthesis quality. As seen in Table~\ref{tab:quanti_celebA_complex}.D-E, the comparison yields similar results to the 2D cases. The trajectories predicted by our method from observations only are more accurate than those inferred by GTM-SM from the ground-truth actions. As to the quality of retrieved and hallucinated images, our method also shows superior performance (\cf additional saliency metrics in the supplementary material). While the current results are still far from visually pleasing, the proposed method is promising, with improvements expected from more powerful generative networks.
Adopting the same metrics as in Section~\ref{sec:exp_2d}, we compare the methods. As seen in Table~\ref{tab:quanti_celebA_complex}.D-E, our method slightly underperforms in terms of localization in the Doom environment. This may be due to the approximate rendering process VizDoom uses for the depth observations, with discretized values not matching the game units. Unlike GTM-SM which relies on action encodings for localization, these unit discrepancies affect our observation-based method.
As to the quality of retrieved and hallucinated images, our method shows superior performance (\cf additional saliency metrics in the supplementary material). While current results are still far from being visually pleasing, the proposed method is promising, with improvements expected from more powerful generative networks.

It should also be noted that the proposed hallucinatory module is more reliable when target scenes have learnable priors (\eg, structure of faces). Hallucination of uncertain content (\eg, layout of a 3D room) can be of lower quality due to the trade-off between representing uncertainties \wrt missing content and unsure localization, and synthesizing detailed (but likely incorrect) images.
Soft registration and hallucinations' statistical nature can add ``uncertainty'' leading to blurred results, which our generative components partially compensate for (\cf our choice of a GAN solution for the DAE to improve its sampling, \cf supplementary material). For data generation use-cases, relaxing hallucination constraints and scaling up $\mathcal{L}_{hallu}$ and $\mathcal{L}_{anam}$ can improve image detail at the price of possible memory corruption (we focused on consistency rather than high-resolution hallucinations).

%-------------------------------------------------------------------------
\section{Conclusion}
\label{sec:cnc}
% !TeX spellcheck = en_US

%We presented a novel framework to perform unobserved scene synthesis. 
Given unlocalized agents only providing observations, our framework builds global representations consistent with the underlying scene properties. Applying prior domain knowledge to harmoniously complete sparse memory, our method can incrementally sample novel views over whole scenes, resulting in the first complete read and write spatial memory for visual imagery. We evaluated on synthetic and real 2D and 3D data, demonstrating the efficacy of the proposed method's memory map. Future work can involve densifying the memory structure and borrowing recent advances in generating high-quality images with GANs \citep{wang2018high}.% More degrees of freedom of the agent movement is also worth exploring.

\pdfbookmark[0]{References}{ref}
\bibliographystyle{named}
\bibliography{references}
% }

%\newpage
\vspace{2em}

\appendix
\beginsupplement

\noindent
\pdfbookmark[0]{Supplementary Material}{sup_mat}
{\Large\textbf{Supplementary Material} \vspace{1em}}

In the following sections, we introduce further pipeline details for reproducibility. We also provide various additional qualitative results (Figures~\ref{fig:home2d_gtmsm_comp2} to \ref{fig:avd_results}) and quantitative comparisons (Section~\ref{sec:supmat_3d}) on 2D and 3D datasets. A video is also attached, presenting our solution applied to the incremental registration of unlocalized observations and generation of novel views.

\section{Methodology and Implementation Details}
% !TeX spellcheck = en_US

This section contains further details regarding the several interlaced components of our pipeline and their implementation.

\subsection{Localization and Memorization}

\subsubsection{Encoding Memories}

Observations are encoded using a shallow ResNet~\cite{he2016identity} with 4 residual blocks (reusing the \textit{CycleGAN} custom implementation of zhis architecture~\cite{zhu2017cyclegan}). The encoder $E$ is thus configured to output feature maps $x'_t \in \mathbb{R}^{n \times h' \times w'}$ with the same dimensions as the inputs $x_t \in \mathbb{R}^{c \times h \times w}$, \ie $h=h', w=w'$.  

As explained in Section~\refwithdefault{sec:encoding}{3.1}, the projection of $x'_t \in \mathbb{R}^{n \times h' \times w'}$ (with features in the image coordinate system) into $o_t \in \mathbb{R}^{n \times s \times s}$, the representation of the agent's spatial neighborhood, is use-case dependent.
For 2D image exploration, this operation is done by cropping $x'_t$ into a square tensor $n \times s' \times s'$ with $s' = min(h',\ w')$, followed by scaling the features from $s' \times s'$ to $s \times s$ using bilinear interpolation.

For 3D use-cases with RGB-D observations, the input depth maps $x_t^d$ are used to project $x'_t$ into a 3D point cloud (after registering color and depth images together), before converting this sparse representation into a dense tensor using max-pooling. For the 3D projection, $\forall i \in \{0, ..., h-1\}$ and $\forall j \in \{0, ..., w-1\}$, each feature $x_{t,i,j} \in \mathbb{R}^n$ of $x'_t$ receives the coordinates $(x,y,z)$ similar to~\cite{henriques2018mapnet}:
\begin{equation}
    z = x^d_{t,i,j} \;\; \text{ ; } \;\;
    x = (j - c_x)\frac{z}{f_x} \;\; \text{ ; } \;\;
    y = (i - c_y)\frac{z}{f_y}
\end{equation}
with $f_x, f_y$ the pixel focal lengths of the depth sensor, and $c_x, c_y$ its pixel focal center (for KinectV2: $f_x = 366.193$px, $f_y = 365.456$px, $c_x = 256.684$px, $c_y = 207.085$px for $512 \times 424$ images).

Each set of coordinates is then discretized to obtain the neighborhood bin the feature belongs to. Given $s \times s$ bins of dimensions $(x_s, z_s)$ in world units, the bin coordinates $(x_b, z_b)$ of each feature are computed as follow:
%\begin{equation}
%    x_b = \floor{\frac{x}{x_s}} + \frac{s-1}{2}\ , \ z_b = \floor{\frac{z}{z_s}} + \frac{s-1}{2}
%\end{equation}
\begin{equation}
    x_b = \floor{\frac{x}{x_s}} + \frac{s-1}{2} \;\; \text{ ; } \;\;
    z_b = \floor{\frac{z}{z_s}} + \frac{s-1}{2}
\end{equation}
with $\floor{\cdot}$ the integer flooring operation. Features projected out of the $s \times s$ area are ignored.

\begin{figure}[t]
\centering
\includegraphics[width=1\linewidth]{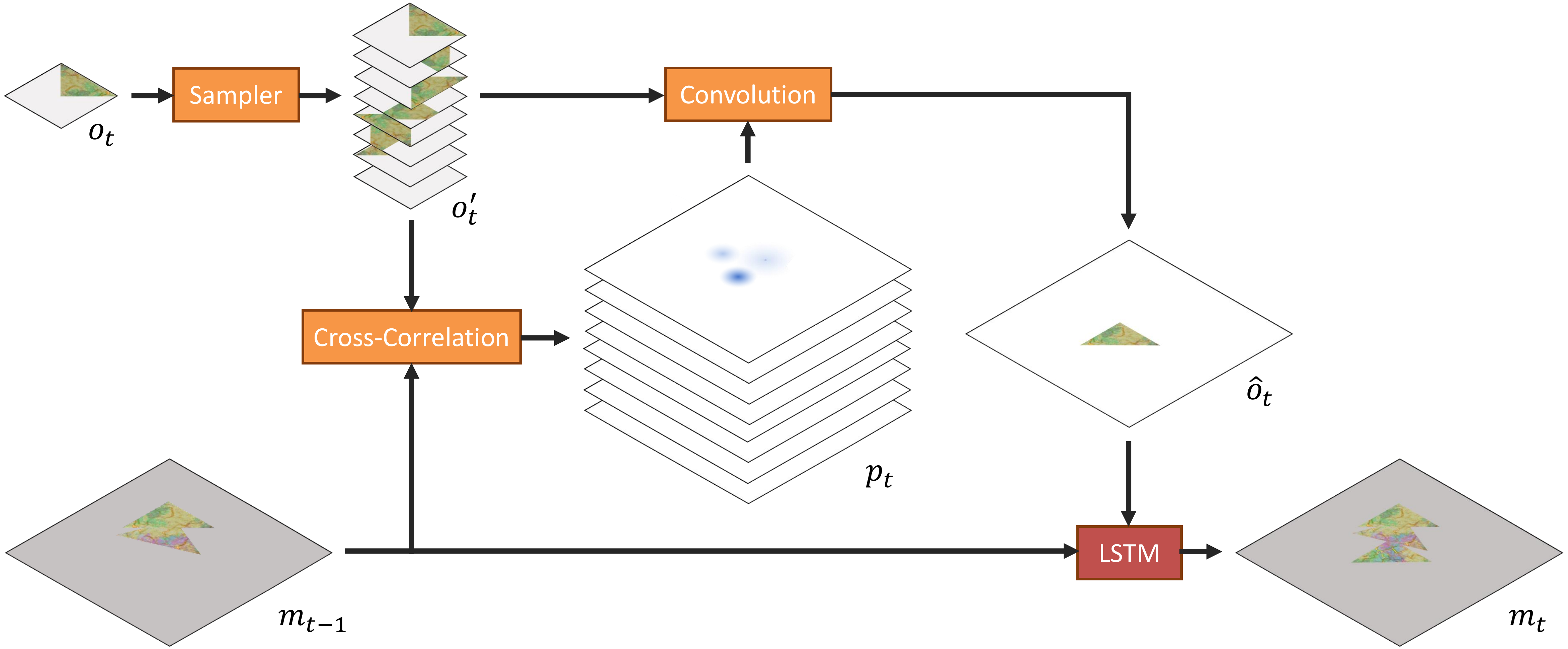}
    \vspace{-1.5em}
\caption{\textbf{Localization and memorization}, based on 
\ MapNet~\cite{henriques2018mapnet}.}
\label{fig:registration}  
\end{figure}

Finally,  $o_t$ is obtained by applying a max-pooling operation over each bin (\ie, keeping only the maximum values for features projected into the same bin). Empty bins result in a null value\footnote{Max-pooling over a sparse tensor (point cloud) as done here is a complex operation not yet covered by all deep learning frameworks at the time of this project. We thus implemented our own (for the PyTorch library~\cite{paszke2017automatic}).}.

\subsubsection{Localizing and Storing Memories}

Once the features are localized and registered into the allocentric system (\cf Figure~\ref{fig:registration}), an LSTM is used to update the global memory accordingly, \cf Section~\refwithdefault{sec:mem-update}{3.1}. Following the original MapNet solution~\cite{henriques2018mapnet}, each spatial location is updated independently to preserve spatial invariance, sharing weights between each LSTM cell. The occupancy mask $b_t$ of the global memory is updated in a similar manner, using an LSTM with shared-weights to update the memory mask with a binary version of $o_t$ (\ie $1$ for bins containing projected features, $0$ otherwise).

\subsection{Anamnesis}

\subsubsection{Memory Culling}

\begin{figure}[t]
\centering
\includegraphics[width=.45\linewidth]{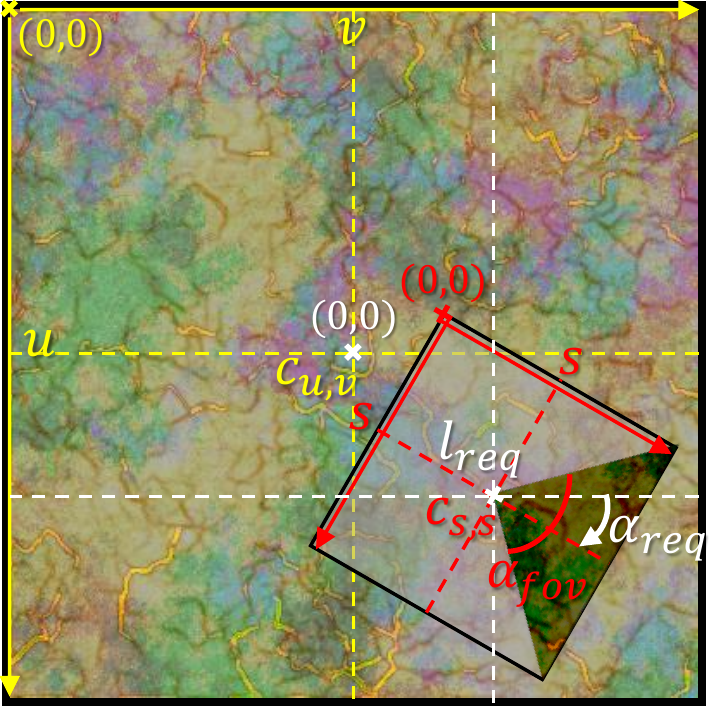}
    \vspace{-.4em}
\caption{\textbf{Geometrical Memory Culling} (2D representation). The feature vector $o_{req}$, defining the observation for an agent positioned at $l_{req}$ and rotated by an angle $\alpha_{req}$ in the allocentric system, is extracted from $m_t$ through a series of geometrical transforms (rotation, translation, clipping, culling). Coordinates and distances in the allocentric coordinate system are represented in white; in yellow for the $m_t$ matrix system; and red for the $o_t$ one.}
\label{fig:culling_geometry}  
\end{figure}

Figure~\ref{fig:culling_geometry} illustrates the geometrical process to extract features from the global map corresponding to the requested viewpoints and agent's view field angle, as described in Section~\refwithdefault{sec:culling}{3.2}.
Note that for this step, one can use a larger view field than requested, in order to provide the feature decoder with more context (\eg, to properly recover visual elements at the limit of the agent's view field).

\subsubsection{Memory Decoding}

Similar to the encoder, we use a ResNet-4 architecture~\cite{he2016identity} for the decoder network, with the last convolutional layers parametrized to output the image tensors $x_{req} \in \mathbb{R}^{c \times h \times w}$ while the network receives inputs feature tensors $o_{req} \in \mathbb{R}^{n \times s \times s}$.

For the experiments where the global memory is directly sampled into an image , another ResNet-4 decoder is trained, directly receiving $m^h_t \in \mathbb{R}^{n \times u \times v}$ for input, returning $x^m_t \in \mathbb{R}^{c \times H \times W}$, and comparing the generated image with the original global image (L1 loss).

\subsection{Mnemonic Hallucination}

For simplicity and homogeneity\footnote{Our solution is orthogonal to the choice of encoding/decoding networks. More advanced architectures could be considered.}, another network based on the ResNet-4 architecture~\cite{he2016identity,zhu2017cyclegan} is used to fill the memory holes. In addition to the global memory $m_t$, the network is adapted to also receive as input a noise vector (which can be seen as a random seed for the hallucinations). This vector goes through two fully-connected layers to upsample it, while the main input goes through the downsampling layers of the ResNet; then the two resulting tensors are concatenated before the first residual block.

In order to improve the sampling of hallucinated features and the global awareness of this generator, we adopt several concepts from SAGAN~\cite{zhang2018self}. The ResNet generator is therefore edited as follow:
\begin{itemize}
    \item Spectral normalization~\cite{miyato2018spectral} is applied to the weights of each convolution layer in the residual blocks (as SAGAN authors demonstrated it can prevent unusual gradients and stabilize training);
    
    \item To model relationships between distant regions, self-attention layers~\cite{zhang2018self,cheng2016long,parikh2016decomposable,vaswani2017attention} replace the two last convolutions of the network.
\end{itemize}
Given a feature map $o \in \mathbb{R}^{n \times u \times v}$, the result $o_{sa}$ of the self-attention operation is:
\begin{equation}
o_{sa} = o + \gamma (W_h \star o) \sigma\big((W_f \star o)^\intercal (W_g \star o)\big)^\intercal
\end{equation}
with $W_f \in \mathbb{R}^{\bar{n} \times n}$, $W_g \in \mathbb{R}^{\bar{n} \times n}$, $W_h \in \mathbb{n}^{n \times n}$ learned weight matrices (we opt for $\bar{n} = \sfrac{n}{8}$ as in~\cite{zhang2018self}); and $\gamma$ a trainable scalar weight.

Following the generative adversarial network (GAN) strategy~\cite{goodfellow2014generative,radford2015unsupervised,salimans2016improved,isola2017pix2pix}, our conditioned generator is also trained against a discriminator evaluating the \textit{realism} of feature patches $o^h_t$ culled from $m^h_t$. This discriminator is itself trained against $o_t$ (\textit{real} samples) and $o^h_t$ (\textit{fake} ones).
For this network, we also opt for the architecture suggested by Zhang~\etal~\cite{zhang2018self}, \ie a simple convolutional architecture with spectral normalization and self-attention layers. 

Given this setup, the generative loss $\mathcal{L}_{hallu}$ is combined to $\mathcal{L}_{disc}$, a discriminative loss obtained by playing the generator $H$ against its discriminator $D$. As a conditional GAN with recurrent elements, the objective this module has to maximize over a complete training sequence is therefore:

\begin{align}
    H^* = {}& \arg \min_H \max_D \mathcal{L}_{disc} + \mathcal{L}_{hallu} + \mathcal{L}_{corrupt} \label{eq:minmax} \\
    \text{with}\;\;
    \mathcal{L}_{disc} = {}& \displaystyle\sum_{t=0}^{\tau}\big[\log D(x_t)\big] + \big[\log \Big(1 - D\big(x^h_t)\Big)\big] \\
\nonumber
\end{align}

\subsection{Further Implementation Details}
	
Our solution is implemented using the PyTorch framework~\cite{paszke2017automatic}.
	
\paragraph{Layer parameterization:}
	\begin{itemize}
		\setlength\itemsep{0em}
		\item Instance normalization is applied inside the ResNet networks;
		\item All Dropout layers have a dropout rate of $50\%$;
		\item All LeakyReLU layers have a leakiness of $0.2$.
		\item Image values are normalized between -1 and 1.
	\end{itemize}
	
\paragraph{Training parameters:}
	\begin{itemize}
		\setlength\itemsep{0em}
		\item Weights are initialized from a zero-centered Gaussian distribution, with a standard deviation of $0.02$ ;
		\item The Adam optimizer~\cite{kingma2014adam} is used, with $\beta_1 = 0.5$;
		\item The base learning rate is initialized at $2e \times 10^{-4}$;
		\item Training sequence applied in this paper:
		    \begin{enumerate}
		        \item Feature encoder and decoder networks are pre-trained together for 10,000 iterations;
		        \item The complete memorization and anamnesis process (encoder, LSTM, decoder) is then trained for 10,000 more iterations;
		        \item The hallucinatory GAN is then added and the complete solution is trained until convergence.
		    \end{enumerate}
	\end{itemize}

\section{Experiments and Results}
% !TeX spellcheck = en_US

Additional results are presented in this section. 
We also provide supplementary information regarding the various experiments we conducted, for reproducibility.

\subsection{Protocol for Experiments}
\label{sec:comp_setup}

\subsubsection{Comparative Setup}

To the best of our knowledge, no other neural method covers agent localization, topographic memorization, scene understanding and relevant novel view synthesis in an end-to-end, integrated manner. The closest state-of-the-art solution to compare with is the recent GTM-SM project~\cite{fraccaro2018generative}. This method uses the differentiable neural dictionary (DND) proposed by Pritzel~\etal~\cite{pritzel2017neural} to store encoded observations with the predicted agent's positions for keys. To synthesize a novel view, the $k$-nearest entries (in terms of positions-keys) are retrieved to interpolate the image features, before passing it to a decoder network.

Unlike our method which localizes and registers together the views with no further context needed, GTM-SM requires an encoding of the agent's actions, leading to each new observation, as additional inputs. We thus adapt our data preparation pipeline for this method, so that the agent returns its actions (encoding the direction changes and step lengths) along the observations. At each time step, GTM-SM uses the provided action $a_t$ to regress the agent's state $s_t$ \ie its relative pose in our experiments. 

For a fair comparison with the ground-truth trajectories, we thus convert the relative pose sequences predicted by GTM-SM into world coordinates. For that, we apply a least-square optimization process to fit its predicted trajectories over the ground-truth ones \ie computing the most favorable transform to apply before comparison (scaling, rotating and translating the trajectories). For our method, the allocentric coordinates are also converted in world units by scaling the values according to the bin dimensions $(x_s, z_s)$ and applying an offset corresponding to the absolute initial pose of the agent.

\begin{figure*}[t]
\centering
\includegraphics[width=1\linewidth]{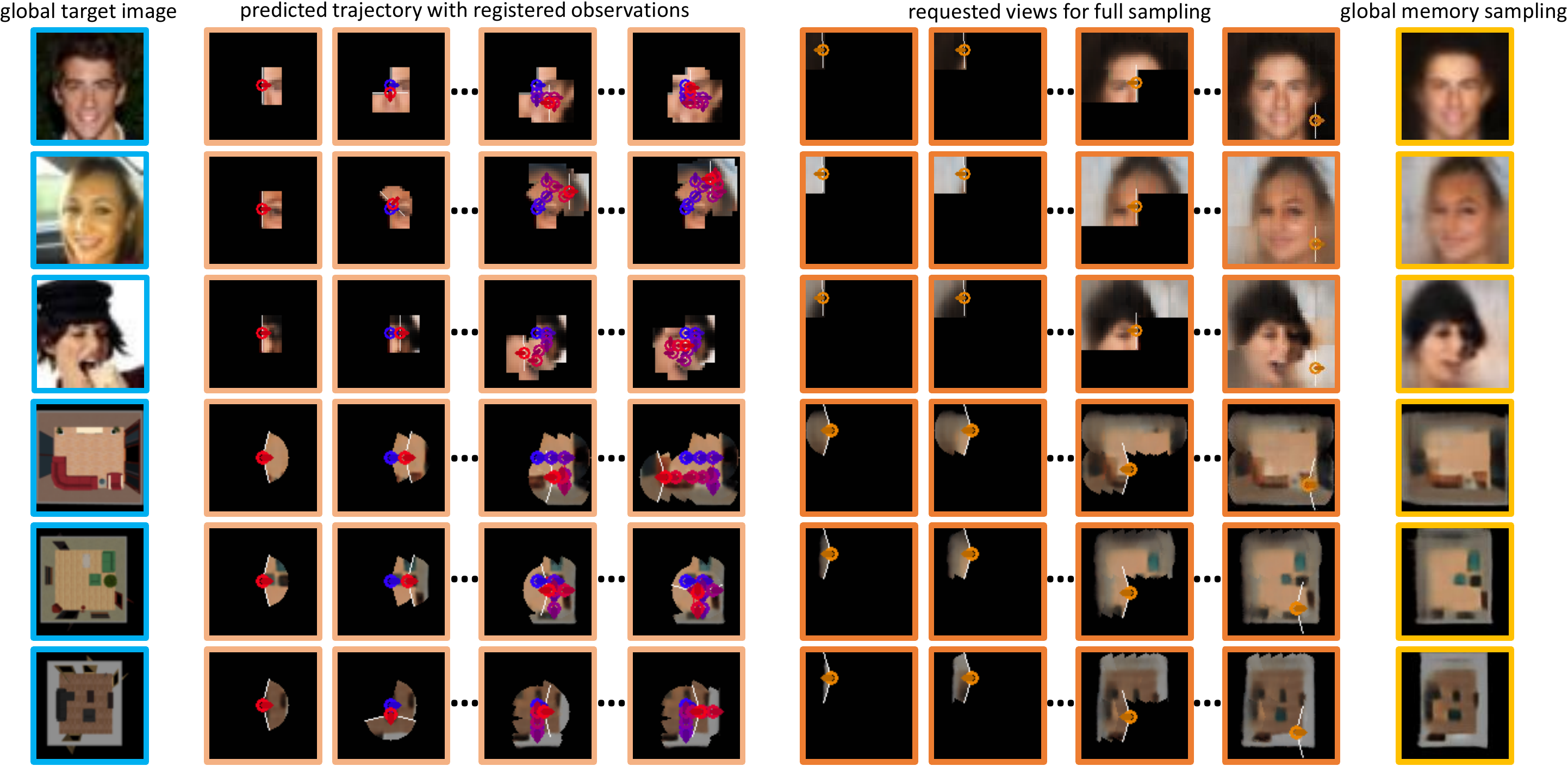}
    \vspace{-1.5em}
\caption{\textbf{Incremental and direct memory sampling} of complete environments from partial observations (on CelebA).
%Our method can iteratively sample complete environments from partial observations (here from CelebA%~\cite{liu2015faceattributes}
%\ and HoME-2D%~\cite{brodeur2017home})
%. The memory can also be directly decoded as a global view (last column).
}
\label{fig:trajectory_full_cover_supmat}  
\end{figure*}

\subsubsection{Metrics for Quantitative Evaluations}
As a reminder (\cf main paper), the following metrics are applied, to evaluate the quality of the localization, the anamnesis, and the hallucination:

\begin{itemize}
    \item The \textit{average position error (APE)} computes the mean Euclidean distance between the predicted positions and their ground-truths for each sequence;
    
    \item The \textit{absolute trajectory error (ATE)} is obtained by calculating the root-mean-squared error in the positions of each sequence, after transforming the predicted trajectory to best fit the ground-truth (giving an advantage to GTM-SM predictions through post-processing, as explained in Section~\ref{sec:comp_setup});
    
    \item The common \textit{L1 distance} is computed as the per-pixel absolute difference between the predicted and expected values, averaged over each image (recalled and/or hallucinated);
    
    \item The \textit{structural similarity ($\SSIM$) index}~\cite{wang2003multiscale,wang2004image}, prevalent in the assessment of perceptual quality~\cite{wang2011information,zhang2012comprehensive,gore2015full}, is computed over $N \times N$ windows extracted from the predicted and ground-truth images, as follows:
    \begin{equation}
      \SSIM(x,\bar{x}) = \frac{(2\mu_x\mu_{\bar{x}} + c_1) + (2 \sigma _{x\bar{x}} + c_2)} 
        {(\mu_x^2 + \mu_{\bar{x}}^2+c_1) (\sigma_x^2 + \sigma_{\bar{x}}^2+c_2)}
    \end{equation}
    with $x$ and $\bar{x}$ the windows extracted from the predicted and ground-truth images, $\mu_x$ and $\mu_{\bar{x}}$ the mean values of the respective windows, $\sigma_x^2$ and $\sigma_{\bar{x}}^2$ their respective variance, $c_1 = 0.001^2$ and $c_2 = 0.003^2$ two constants for numerical stability. The final index is computed by averaging the values obtained by sliding the windows over the whole images (no overlapping). We opt for $N=5$ for the CelebA experiments, and $N=13$ for the HoME-2D ones (\ie, splitting the observations in 9 windows). Note that the closer to $1$ the computed index, the better the perceived image quality.

\end{itemize}

\subsection{Navigation in 2D Images}

\subsubsection{Experimental Details}

The aligned CelebA dataset~\cite{liu2015faceattributes} is split with $197,599$ real portrait images for training and $5,000$ for testing. Each image is center-cropped to $160 \times 160$px (in order to remove part of the background and focus on faces), before being scaled to $43 \times 43$px.

We build our synthetic HoME-2D dataset by rendering several thousand RGB floor plans of randomly instantiated rooms using the HoME framework~\cite{brodeur2017home} (room categories: ``\textit{bedroom}'', ``\textit{living}'', ``\textit{office}'') and SUNCG data \cite{SUNCG}. We use $8,960$ images for training and $2,240$ for testing, scaled to $83 \times 83$px.

For both experiments, sequences of observations are generated by randomly walking an agent over the 2D images. At each step, the agent can rotate maximum $\pm 90^\circ$ (for experiments with rotation) and cover a distance from $\sfrac{1}{4}$ to $\sfrac{3}{4}$ its view field radius. However, once a new direction is chosen, the agent has to take at least 3 steps before being able to rotate again (to favor exploration). The agent is also forced to rotate when one of the image borders is entering its view field.

Each training sequence contains $54$ images for the CelebA experiments, and $41$ for the HoME experiments. Both for GTM-SM~\cite{fraccaro2018generative} and our method, only 10 images are passed as observations to fill and train the topographic memory systems (using the provided ground-truth positions/orientations). The remaining 44 or 31 images (sampled by forcing the agent to follow a pre-determined trajectory covering the complete 2D environments) are used as ground-truth information for the hallucinatory modules of the two pipelines.

As described in Section~\refwithdefault{sec:exp_2d}{4.1}, for each dataset we consider two different types of agents, \ie with more or less realistic characteristics:

\paragraph{Simple agent.} We first consider a non-rotating agent---only able to translate in the four directions---with a $360^\circ$ view field covering an image patch centered on the agent's position. For CelebA experiments, this view field is $15 \times 15$px square patch; while for HoME-2D experiments, the view field reaches $20$px away from the agent, and is therefore in the shape of a circular sector (pixels in the corresponding $41 \times 41$ patches further than $20$px are set to the null value).

\paragraph{Advanced agent.} A more realistic agent is also designed, able to rotate and to translate accordingly (\ie in the gaze direction) at each step, observing the image patch in front of it (rotated accordingly). For CelebA experiments, the agent can rotate by $\pm 45^\circ$ or $\pm 90^\circ$ each step, and only observes the $8 \times 15$ patches in front ($180^\circ$ rectangular view field); while for  HoME-2D experiments, it can rotate by $\pm 90^\circ$ each step, and has a $150^\circ$ view field limited to $20$px.

The first simple agent is defined to reproduce the 2D experiments showcasing GTM-SM~\cite{fraccaro2018generative}. While its authors present some qualitative evaluation with a rotating agent, we were not able to fully reproduce their results, despite the implementation changes we made to take into account the prior dynamics of the moving agent (\ie extending the GTM-SM state space to 3 dimensions; the new third component of the state vectors $s_{t-1}$ is storing the information to build a 2D rotation matrix, itself used with the translation elements to compute $s_t$). We adopt the more realistic agent to demonstrate the capability of our own solution, given its more complex range of actions and partial observations. Rotational errors with this agent are ignored for GTM-SM.

\subsubsection{Additional Qualitative Results}

\begin{figure*}[ht]
\centering
\includegraphics[width=\linewidth]{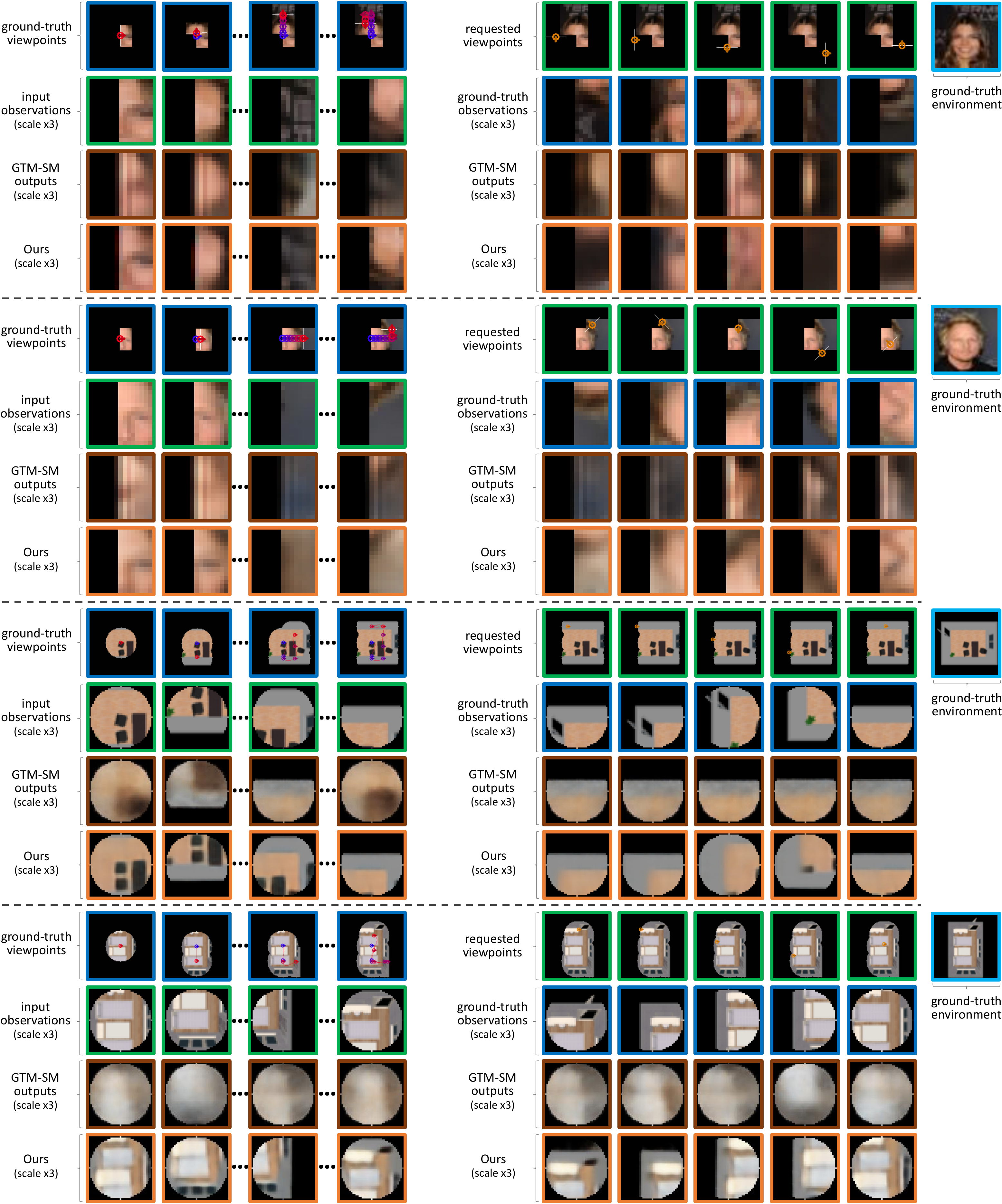}
    \vspace{-1.5em}
\caption{\textbf{Qualitative comparison with GTM-SM} on CelebA and HoME-2D, in terms of pose / trajectory estimations and in terms of view generation (recovery of seen images from memory and novel view hallucination). Methods receive a sequence of 10 observations (along with the related actions for GTM-SM) from a non-rotating agent exploring the $83 \times 83$ 2D image with a 360$^\circ$ circular view field of $20$px radius. The methods then apply their knowledge to generate novel views.}
\label{fig:home2d_gtmsm_comp2}  
\end{figure*}

\begin{figure*}[ht]
\centering
\includegraphics[width=\linewidth]{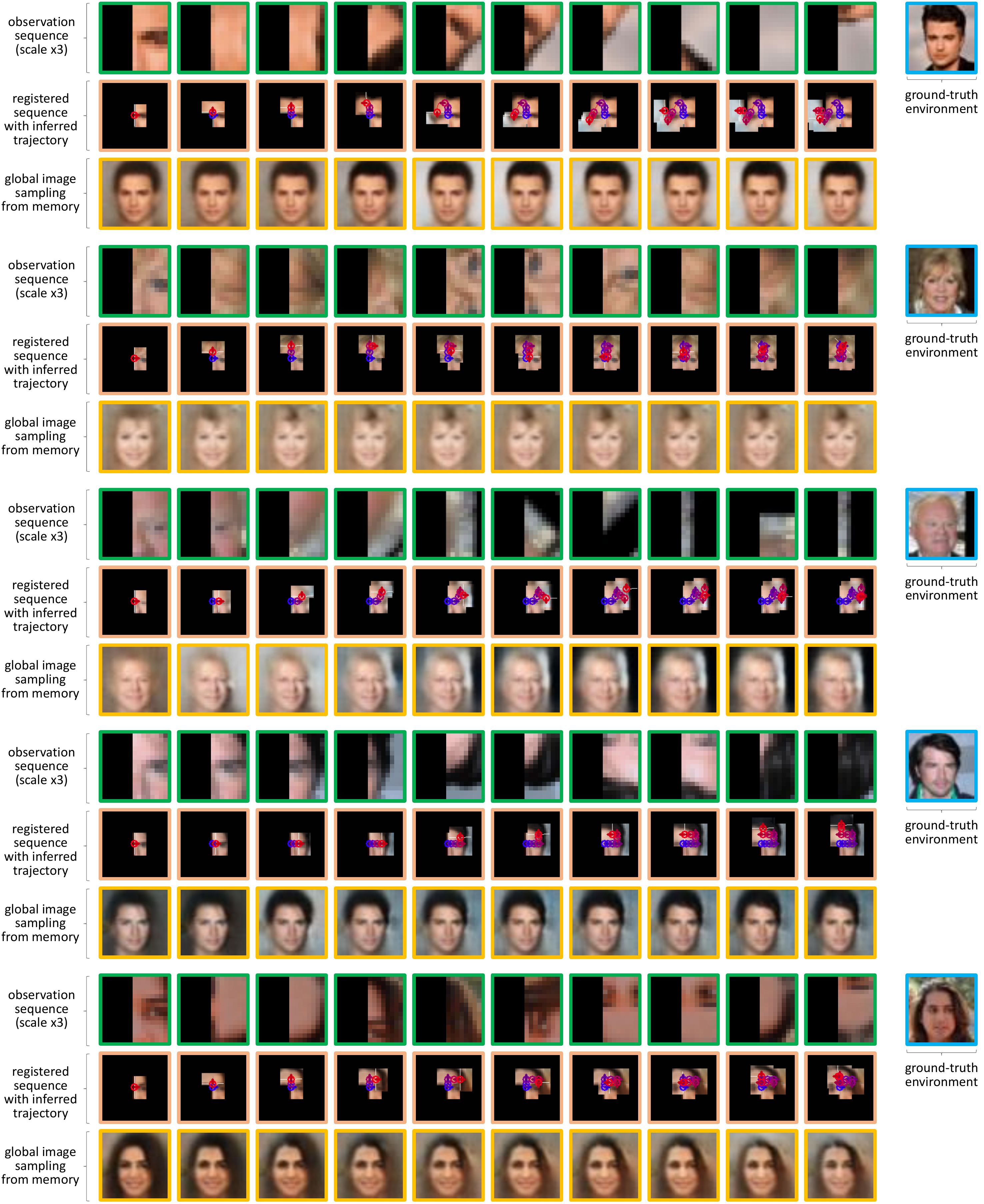}
    \vspace{-1.5em}
\caption{\textbf{Qualitative results on CelebA dataset}. with sequences of 10 observations from an agent able to rotate and translate every step, exploring the $43 \times 43$ 2D image with a 180$^\circ$ view field of $8 \times 15$px (\ie observing the image patch in front of it, rotated accordingly). After each step, the hallucinated features are adapted to blend with the new observations, until reaching convergence as the coverage increases.}
\label{fig:celeba_supmat_grid}  
\end{figure*}

\begin{figure*}[ht]
\centering
\includegraphics[width=\linewidth]{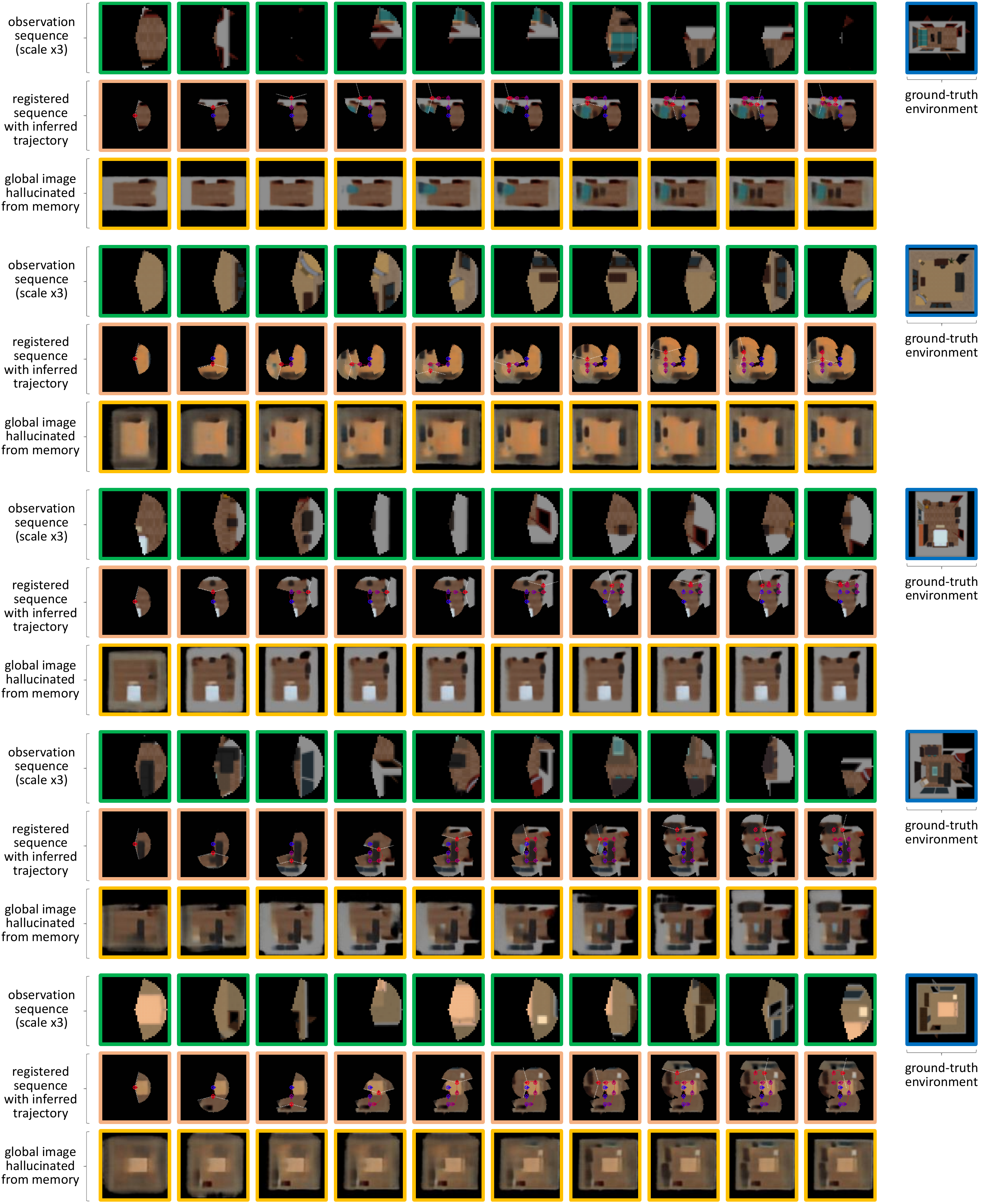}
    \vspace{-1.5em}
\caption{\textbf{Qualitative results on HoME-2D}. with sequences of 10 observations from an agent able to rotate and translate every step, exploring the $83 \times 83$ 2D image with a 150$^\circ$ view field of $20$px radius. After each step, the hallucinated features are adapted to blend with the new observations, until reaching convergence as the coverage increases.}
\label{fig:home2d_supmat_grid}  
\end{figure*}

As explained in the paper, our topographic memory module not only allows to directly build a global representation of the environments, but it also brings the possibility to use prior knowledge to extrapolate the scene content for the unexplored area. In contrast, GTM-SM stores each observation separately in its DND memory~\cite{pritzel2017neural}, and can only generate new views by interpolating between a subset of these entries with a VAE prior. This is illustrated in Figure~\refwithdefault{fig:celeba_gtmsm_comp2}{4} (comparing the methods on image retrieval from memory and on novel view synthesis) and Figure~\refwithdefault{fig:celebahome_incremental_hallu}{6} (showcasing the ability of our pipeline to synthesize complete environments from partial views) in the main paper, as well as in similar
Figures~\ref{fig:trajectory_full_cover_supmat} to \ref{fig:home2d_supmat_grid}.
%Figures~\ref{fig:celeba_gtmsm_comp2}-\ref{fig:home2d_gtmsm_comp2} and \ref{fig:celeba_supmat_grid}-\ref{fig:home2d_supmat_grid}.

%We invite our readers to also have a look at a web-hosted video which presents further qualitative results. The video can be found on this web page: \url{https://sites.google.com/view/incremental-scene-synthesis}. Given various agents (with different behaviors, view fields, sequence lengths, \etc) walking on top of images from CelebA~\cite{liu2015faceattributes} and HoME-2D~\cite{brodeur2017home}, we demonstrate how our pipeline can both properly register the partial observations in its global memory (inferring the agent's trajectory coincidentally) and synthesize new relevant and consistent views. This process can be applied to the incremental generation of images covering the complete environments, from observed features and prior domain knowledge.

\subsection{Exploring Real 3D Scenes}
\label{sec:supmat_3d}

\subsubsection{Additional Quantitative Results}

Additional evaluations were conducted on real 3D data, using the Active Vision Dataset (AVD)~\cite{active-vision-dataset2017}, in order to demonstrate the salient properties of the generated images (despite their lower visual quality).

First, the Wasserstein metric was computed between the Histogram of Oriented Gradients (HOG) descriptors extracted from the unseen ground-truth images and the corresponding predictions. GTM-SM~\cite{fraccaro2018generative} scored 1.1, whereas our method obtained 0.8 (the lower the better).

Second, we compared the saliency maps of ground-truth and predicted images~\cite{salMetrics_Bylinskii}, computing the area-under-the-curve metric (AUC) proposed by Judd \etal~\cite{judd2009learning} and the Normalized Scanpath Saliency (NSS)~\cite{peters2005components}. GTM-SM~\cite{fraccaro2018generative} scored 0.40 for the AUC-Judd and 0.14 for the NSS, whereas our framework obtained 0.63 and 0.38 respectively (the higher the better for both metrics).

\subsubsection{Additional Qualitative Results}
% \cBen{Add: figure comparing views and predicted trajectories for our method and GTM-SM + opt. M2N on AVD - at least 3 sequences \ie Figure~\ref{fig:avd_results}}

% \cBen{Explain properly: Complicated task. But our method manages to build a 3D representation of the scene and generate novel views accordingly, properly rendering some of the room features. On the other hand, GTM-SM fails.}

For qualitative comparison with GTM-SM, we trained both methods on AVD dataset with the same setup. Challenges arise from the fact that the 3D environments are much more complex than their 2D counterparts, and more factors need to be considered in memorization and prediction.

\begin{figure*}[ht]
\centering
\includegraphics[width=\linewidth]{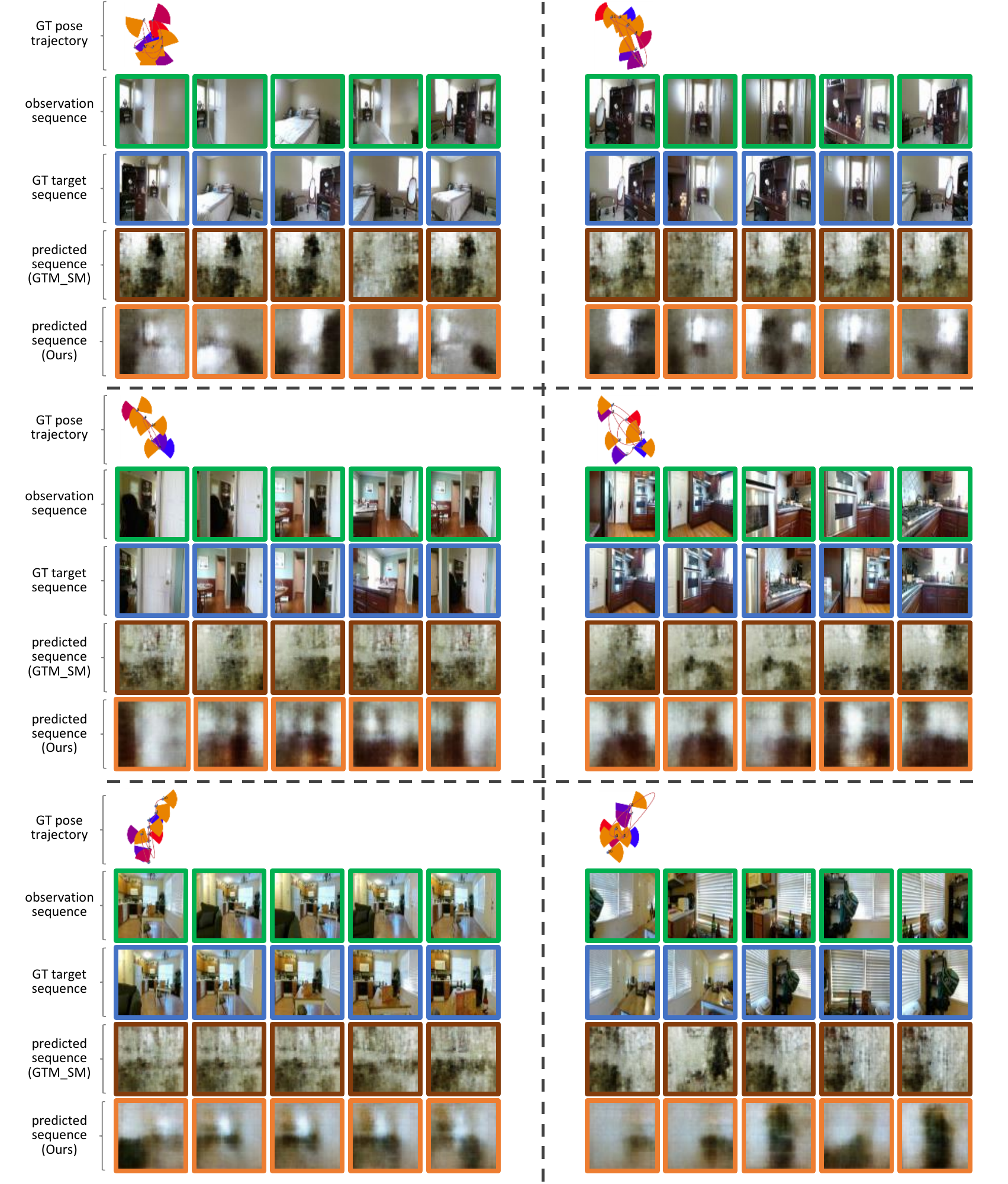}
    \vspace{-1.5em}
\caption{\textbf{Qualitative comparison with GTM-SM on AVD dataset}, in terms of pose / trajectory estimations and in terms of view generation (recovery of seen images from memory and novel view hallucination). Methods receive a sequence of 5 observations (along with the corresponding actions for GTM-SM) from an agent exploring the testing unseen scenes. The methods then apply their knowledge to generate novel views.}
\label{fig:avd_results}  
\end{figure*}

Further qualitative results on the AVD test scenes are demonstrated in Figure~\ref{fig:avd_results}. Given the same observation sequences (additional actions to GTM-SM) and requested poses, the predicted novel views are shown for comparison. Generally, GTM-SM fails to adapt the VAE prior and predict the belief of target sequences, while our method tends to successfully synthesize the room layout based on the learned scene prior and observed images.

%\begin{figure*}[ht]
%\centering
%\includegraphics[width=\linewidth]{figures/celeba_gtmsm_comp2.pdf}
%\caption{\textbf{Qualitative comparison with %GTM-SM~\cite{fraccaro2018generative} on CelebA %dataset}~\cite{liu2015faceattributes}. Methods receive a sequence of 10 %observations (along with the related actions for GTM-SM) from an agent, %able to rotate $\pm 45^\circ$ or $\pm 90^\circ$ every step and exploring %the $43 \times 43$ 2D image with a 180$^\circ$ view field of $8 \times %15$px (\ie observing the image patch in front of it, rotated accordingly). %The methods then apply their knowledge to generate novel views.}
%\label{fig:celeba_gtmsm_comp2}  
%\end{figure*}

%\begin{figure*}[!ht]
%\centering
%\includegraphics[width=\linewidth]{figures/avd2.pdf}
%\caption{\textbf{Sample qualitative results on 3D realistic Active Vision Dataset~\cite{active-vision-dataset2017}}. Given 5 observations, our solution recovers the 3D topography of the scene, which allows the sampling of relevant novel views.}
%\label{fig:avd_quali}  
%\end{figure*}

\end{document}